\documentclass[10pt,journal,compsoc]{IEEEtran}

\ifCLASSOPTIONcompsoc
  \usepackage[nocompress]{cite}
\else
  \usepackage{cite}
\fi

\usepackage{amsmath,amssymb,amsfonts}
\usepackage{algorithmic}
\usepackage{graphicx}
\usepackage{textcomp}
\usepackage{multirow}
\usepackage{xcolor}
\usepackage{times}
\usepackage{framed}
\usepackage[ruled,vlined,linesnumbered]{algorithm2e}
\usepackage[flushleft]{threeparttable}
\usepackage{graphicx}
\usepackage {subcaption}
\usepackage{url}
\usepackage{subcaption}
\usepackage{makecell}
\newcommand{\cmmnt}[1]{}{}

\makeatletter
\newcommand*{\rom}[1]{\expandafter\@slowromancap\romannumeral #1@}
\newcommand{\tabincell}[2]{\begin{tabular}{@{}#1@{}}#2\end{tabular}}
\makeatother

\newtheorem{example}{Example}

\begin{document}

\title{Repairing Adversarial Texts through Perturbation}

\author{
 Guoliang Dong, Jingyi Wang, Jun Sun, Sudipta Chattopadhyay,\\Xinyu Wang, Ting Dai, Jie Shi and Jin Song Dong
   \IEEEcompsocitemizethanks{
           \IEEEcompsocthanksitem Guoliang Dong, Jingyi Wang and Xinyu Wang are with the Zhejiang University. \protect\\
            E-mail: \{dgl-prc, wangxinyu\}@zju.edu.cn, wangjyee@gmail.com
            \IEEEcompsocthanksitem Jun Sun is with the Singapore Management University.\protect\\
            E-mail: junsun@smu.edu.sg
            \IEEEcompsocthanksitem  Sudipta Chattopadhyay is with Singapore University of Technology and Design. \protect\\
            E-mail: sudipta\_chattopadhyay@sutd.edu.sg
            \IEEEcompsocthanksitem  Ting Dai and Jie Shi are with Huawei International Pte. Ltd. \protect\\
            E-mail: \{daiting2, SHI.JIE1\}@huawei.com
            \IEEEcompsocthanksitem  Jin Song Dong is with the National University of Singapore. \protect\\
            E-mail: dongjs@comp.nus.edu.sg            
}
}
% \markboth{Journal of \LaTeX\ Class Files,~Vol.~14, No.~8, August~2015}%
% {Shell \MakeLowercase{\textit{et al.}}: Bare Demo of IEEEtran.cls for Computer Society Journals}
\IEEEtitleabstractindextext{%
\begin{abstract}
It is known that neural networks are subject to attacks through adversarial
perturbations, i.e., inputs which are maliciously crafted through perturbations to
induce wrong predictions. Furthermore, such attacks are impossible to eliminate,
i.e., the adversarial perturbation is still possible after applying mitigation
methods such as adversarial training. Multiple approaches have been developed to
detect and reject such adversarial inputs, mostly in the image domain. Rejecting
suspicious inputs however may not be always feasible or ideal. First, normal
inputs may be rejected due to false alarms generated by the detection algorithm.
Second, denial-of-service attacks may be conducted by feeding such systems with
adversarial inputs. To address the gap, in this work, we propose an approach to
automatically repair adversarial texts at runtime. Given a text which is
suspected to be adversarial, we novelly apply multiple adversarial perturbation
methods in a positive way to identify a repair, i.e., a slightly mutated but
semantically equivalent text that the neural network correctly classifies. Our
approach has been experimented with multiple models trained for natural language
processing tasks and the results show that our approach is effective, i.e., it
successfully repairs about 80\% of the adversarial texts. Furthermore, depending
on the applied perturbation method, an adversarial text could be repaired in as
short as one second on average.
\end{abstract}
\begin{IEEEkeywords}
Adversarial Text, Detection, Repair, Perturbation.
\end{IEEEkeywords}}

\maketitle

\section{Introduction}

Neural networks (NNs) have achieved state-of-the-art performance in many tasks,
such as classification, regression and planning~\cite{rastegari2016xnor,
xu2015regression,segler2018planning}. For instance, text classification is one
of the fundamental tasks in natural language processing (NLP) and has broad
applications including sentiment analysis~\cite{dos2014deep, tang2015document},
spam detection~\cite{ren2017neural, jain2019spam} and topic
labeling~\cite{yang2016hierarchical}. NNs have been shown to be effective in
many of these text classification tasks~\cite{zhang2015character}.

\begin{figure}
	\centering
	\includegraphics[width=0.4\textwidth]{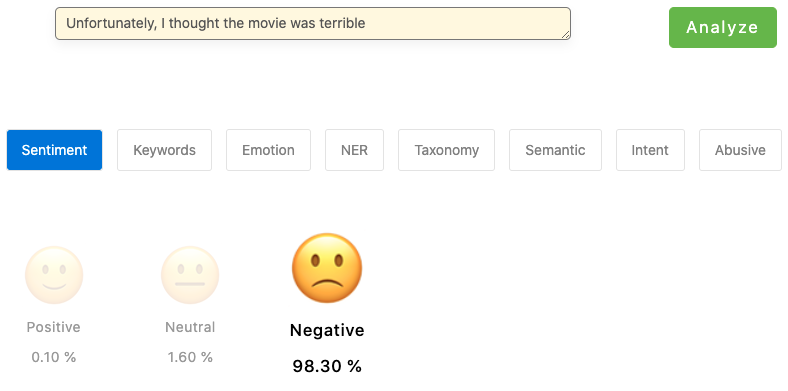}
	\includegraphics[width=0.4\textwidth]{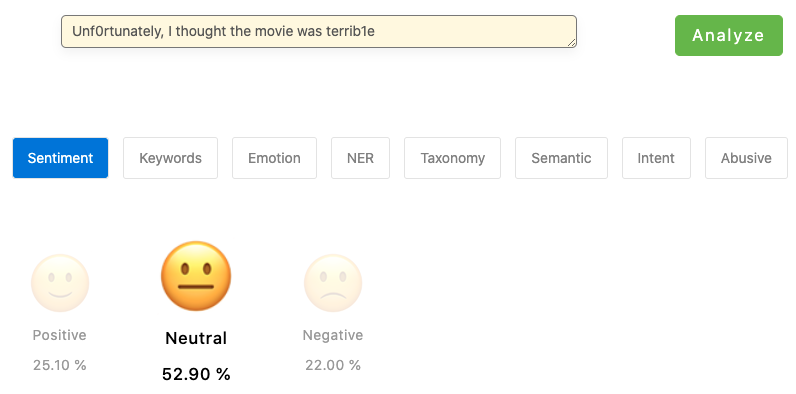}
	\caption{Example of adversarial text on ParallelDots (a sentiment analysis API). The upper is the result of the original
	 text and the lower is that of the adversarial text.}
	\label{fig:pd}
\end{figure}

At the same time, NNs are found to be vulnerable to various attacks, which raise
many security concerns especially when they are applied in safety-critical
applications. In particular, it is now known that NNs are subject to adversarial
perturbations~\cite{szegedy2013intriguing}, i.e., a slightly modified input may
cause an NN to make a wrong prediction. Many attacking methods have been
proposed to compromise NNs designed and trained for a variety of application
domains, including images~\cite{goodfellow2014explaining, carlini2017towards},
audio~\cite{carlini2018audio} and texts~\cite{hotflip,
papernot2016crafting,samanta2017towards}. Multiple approaches like
HotFlip~\cite{hotflip} and TEXTBUGGER~\cite{textbugger} have been proposed to
attack NNs trained for text classification. TEXTBUGGER attacks by identifying
and changing certain important characters (or words) in the text to cause a
change in the classification result. For example, given the text
\emph{``\textit{Unfortunately}, I thought the movie was \textit{terrible}''}
which is classified as `negative' by an NN for sentiment analysis, TEXTBUGGER
produces an adversarial text \emph{``\textit{Unf0rtunately}, I thought the movie
was \textit{terrib1e}''} which is classified as `neutral', as shown in Fig~\ref{fig:pd}. While the above
perturbation is detectable with a spell checker, there are also attacking
methods like SEAs~\cite{sears} which generate adversarial texts that are hard to
detect.

Efforts on defending against adversarial attacks fall into two categories. One is to
train a robust classifier which either improves the accuracy on such examples,
e.g., adversarial training~\cite{madry2017towards, ShafahiNG0DSDTG19} and
training models with pre-processing samples generated by dimensionality reduction or JPEG
compression~\cite{pcadefense, das2017keeping}, or decreases the success rate for attackers on
generating adversarial samples, e.g., obfuscated gradients~\cite{Guo@2018,
Dhillon18}. None of these approaches, however, can eliminate adversarial samples completely~\cite{Anish@2018} as the
adversarial samples may not be a flaw of the model but features in
data~\cite{Andrew19}. Alternative mitigation approaches alleviate 
the effects of such samples by detecting adversarial
samples~\cite{zheng2018robust,liu2019detection,xu2019adversarial,wang2019adversarial}.

Although most of the detecting approaches have focused on the image domain,
simple approaches have been proposed to detect adversarial texts as well. One
example is to apply a character/word checker, i.e., Gao \emph{et
al.}~\cite{gao2018black}, to detect adversarial texts generated by
HotFlip~\cite{hotflip} and TEXTBUGGER~\cite{textbugger}. Detecting adversarial
samples is however not the end of the story. The natural follow-up question is
then: what do we do when a sample is deemed adversarial? Some approaches simply
reject those adversarial samples~\cite{lu2017safetynet,meng2017magnet,wang2019adversarial}. Rejection is
however not always feasible or ideal. First, existing detection algorithms often
generate a non-negligible amount of false alarms~\cite{xu2019adversarial,wang2019adversarial}, particularly so for the
simple detection algorithms proposed for adversarial
texts~\cite{rosenberg2019defense}. Second, rejection may not be an option for certain applications. For example, it is not
a good idea to reject an edit in public platforms (e.g., Wikipedia, Twitter and
GitHub) even if the edit is suspected to be maliciously crafted (e.g.,
toxic)~\cite{hosseini2017deceiving}. Rather, it would be much better to suggest a 
minor ``correction'' on the edit so that it is no longer malicious. Lastly,
rejecting all suspicious samples would easily lead to deny-of-service attacks.

Beyond rejection, in the image domain, a variety of techniques are proposed to mitigate the effect of the adversarial samples after these samples are
identified. For example, Pixeldefend~\cite{song2017pixeldefend} rectified the suspicious input images by changing them slightly towards the training
distribution. Akhtar \emph{et al.}~\cite{akhtar2018defense} attached a network
to the first layer of the target NN to reconstruct clean images from the
suspicious ones. Agarwal \emph{et al.}~\cite{agarwal2020image} proposed to use wavelet transformation
and inverse wavelet to remove the adversarial noise. Besides, Goswami
\emph{et al.}~\cite{goswami2019detecting,goswami2018unravelling} proposed a
selective dropout method which mitigates the problem of adversarial samples by
removing the most problematic filters of the target NN.

However, to the best of our knowledge, the question that whether we can
effectively repair adversarial \textit{texts} has been largely overlooked so
far. Even worse, the aforementioned mitigation approaches can not be easily
extended to the text domain due to several fundamental new challenges. To
address the gap, in this work, we aim to develop an approach that automatically
repairs adversarial texts. That is, given an input text, we first check whether
it is adversarial or not. If it is deemed to be adversarial, we identify a
slightly mutated but semantically equivalent text which the neural network
correctly classifies as the suggested repair.

Two non-trivial technical questions must be answered in order to achieve our
goal. \textit{First, how do we generate slightly mutated but semantically equivalent
texts?} Our answer is to novelly apply adversarial perturbation methods in a
positive way. One of such methods is the SEAs attacking method which generates
semantically `equivalent' texts by applying neural machine translation (NMT)
twice (i.e., translate the given text into a different language and back).
Another example is a perturbation method which is developed based on TEXTBUGGER,
i.e., identifying and replacing important words in a sentence with their synonyms.
\textit{Second, how do we know what is the correct label, in the presence of adversarial
texts?} 

Our answer is differential testing combined with majority voting. Given two or more
NNs trained for the same task, our intuition is that if there is a disagreement
between the NNs, the labels generated by the NNs are not reliable. Due to the
transferability of adversarial samples~\cite{goodfellow2014explaining}, a
label agreed upon by the NNs may still not be reliable. We thus propose to
compare the outputs of the models (in the form of probability vectors) based on
KL divergence~\cite{joyce2011kullback} to identify the correct label.
Furthermore, we apply the sequential probability ratio test (SPRT) algorithm to systematically
evaluate the confidence of each possible label and output the
most-likely-correct label based on majority voting only if it reaches certain level of
statistical confidence. 

We implemented our approach as a self-contained toolkit targeting NNs trained
for text classification tasks. Our experiments on multiple real-world tasks
(e.g., for sentiment analysis and topic labeling) show that our approach can
effectively and efficiently repair adversarial texts generated using two
state-of-art attacking methods. In particular, we successfully repair about 80\%
of the adversarial texts, and depending on the applied perturbation method, an
adversarial text could be repaired in as few as 1 second on average.

In summary, we make the following main contributions.
\begin{itemize}
\item We propose the first approach to repair adversarial texts.
	\item We propose, as a part of our overall approach, an approach for detecting adversarial texts based on an enhanced variant of differential testing.
	\item We implement a software toolkit and evaluate it on multiple state-of-the-set NLP tasks and models.
\end{itemize}

The rest of the paper is organized as follows. In Section~\ref{sec:back}, we present relevant background. In Section~\ref{sec:app}, the details of our approach are presented. Section~\ref{sec:exp} shows our experimental setup and results. We discuss related works in Section~\ref{sec:re} and conclude in Section~\ref{sec:con}.

 \section{Background}
\label{sec:back}
In this section, we present a background that is relevant to this work.

\subsection{Text classification}
\emph{Text classification} is one of the most common tasks in Natural Language
Processing (NLP). The objective is to assign one or several pre-defined labels
to a text. Text classification is widely applied in many applications such as
sentiment analysis~\cite{dos2014deep, tang2015document}, topic
detection~\cite{yang2016hierarchical} and spam detection~\cite{ren2017neural,
jain2019spam}. Neural networks (NNs) have been widely adopted in solving text
classification tasks. In particular, Recurrent Neural Networks (RNNs, e.g.,
LSTM~\cite{lstm} and GRU~\cite{gru}), designed to deal with sequential data, are
commonly applied in many NLP tasks. In addition, Convolutional Neural Networks
(CNNs) are shown to achieve similar results on text classification
tasks~\cite{textcnn}. In this work, we focus on RNNs and CNNs and leave the
evaluation of other models to future work.

\subsection{Generating Adversarial Texts}
In the following paragraphs, we introduce state-of-the-art approaches to
generate adversarial texts for NNs.\\

\emph{HotFlip}. HotFlip is a white-box attacking method that generates
adversarial texts at the character level~\cite{hotflip}. Given an input, HotFlip
first finds the best position for attacking the text according to the
directional derivatives and then performs one of the three operations at the
identified position, i.e., substitute one character, insert a character, or
delete the character. Note that these operations usually lead to a meaningless
word and as a consequence, such attacks are easily detected by a spell-checker. 

\emph{TEXTBUGGER}. TEXTBUGGER is a general framework for crafting adversarial
texts~\cite{textbugger}. Given an input text, it first identifies the most
important sentence and then the most important word. A word is the most
important if changing it leads to the most decrease in the classification
confidence. After a word is selected, five operations are applied to generate
adversarial texts. Four of the five operations, i.e., inserting a space in the
word, deleting a character, swapping two characters and substituting a character
with a similar one (like ``o'' to ``0''), are character-level operations which
aim to generate ``human-imperceptible'' texts. Similarly, adversarial texts
generated by these operations are easily detected by a spell-checker. The last
operation is to substitute the selected word with a synonym (hereafter
\textit{Sub-W}), which is hard to detect and likely semantic-preserving.

\emph{TEXTFOOLER}. TEXTFOOLER is another recent method to generate
adversarial texts~\cite{textfooler}. Instead of sorting sentences by importance
at first, TEXTFOOLER directly performs word importance ranking, and then replaces
the words in the ranking list one by one with a synonym until the prediction of
the target model is changed. In general, HotFlip, TEXTBUGGER, and TEXTFOOLER
share the same idea of crafting adversarial texts, and they mainly differ in the substitution of selected words. Among the three methods,
TEXTFOOLER is more likely to generate more natural adversarial texts since it
takes the part of speech into account when selecting synonyms.

\emph{SEAs}. SEAs aims to generate semantic-equivalent adversarial texts by paraphrasing~\cite{sears} based on Neural Machine Translation (NMT).
NMT is a category of NNs which are trained for machine translation and has achieved state-of-the-art performance in machine translation~\cite{kalchbrenner2013recurrent,sutskever2014sequence,cho2014properties}.
SEAs applies NMTs to generating semantic-preserving texts as adversarial texts. That is, SEAs translates an input sentence into multiple foreign languages and then translates them back to the source language using NMTs. After that, SEAs selects an adversarial text among those according to a semantic score, which measures how semantic-preserving is the text with respect to the original input.

\begin{figure*}[t]
    \centering
    \includegraphics[width=0.7\textwidth]{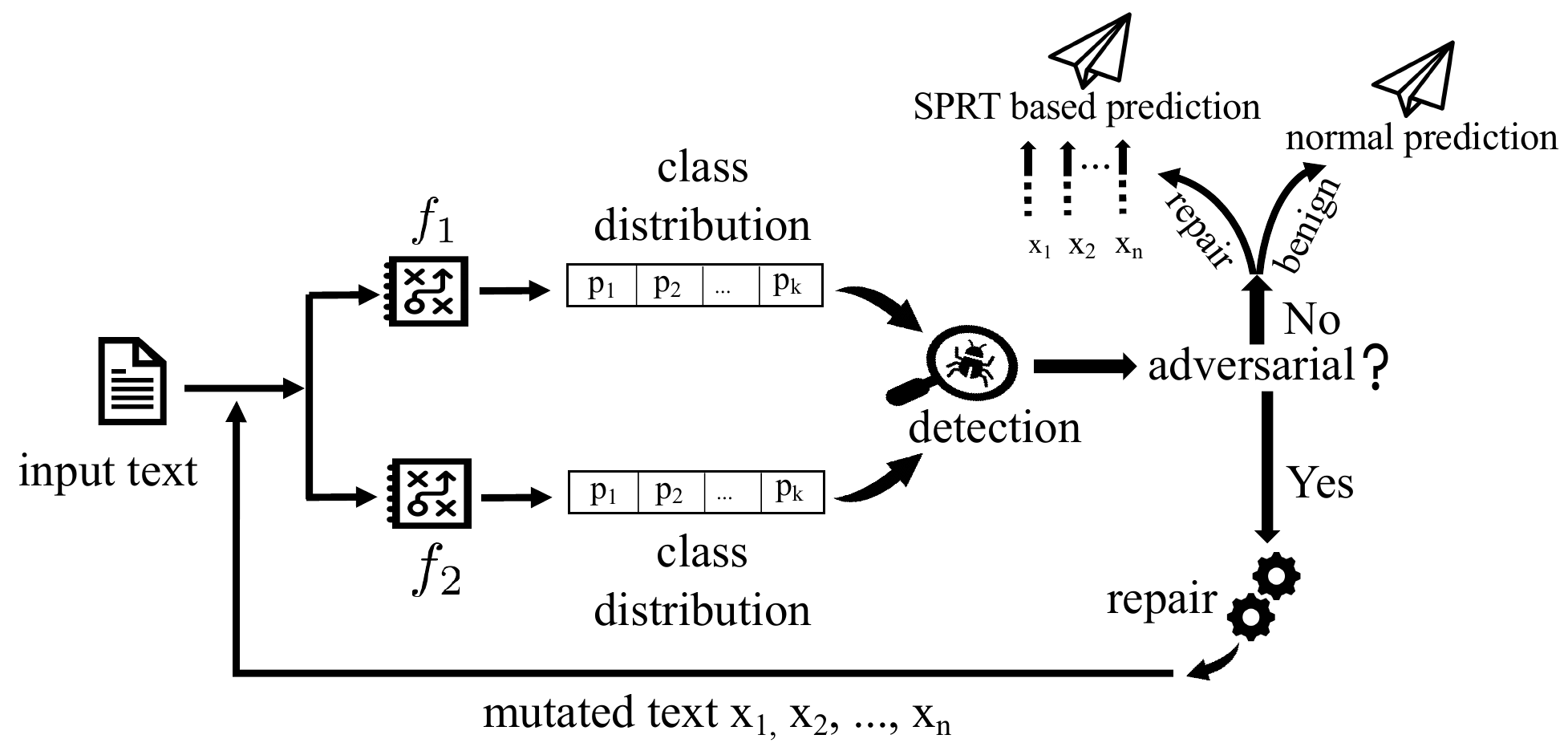}
    \caption{Framework of our approach. Given an input text, we first check if
    it is adversarial with two models $f_1$ anf $f_2$, and then we continually
    generate mutated text to restore its true label with SPRT if it is
    identified as adversarial.}
    \label{fig:framework}
\end{figure*}
\section{Our Repair Approach}
\label{sec:app}

Our aim is to automatically repair adversarial texts. We define our problem as
follows. Given a text input $x$ and a pair of different NNs $(f_1,f_2)$ which
are trained for the same task, how to automatically check whether $x$ is likely
adversarial and generate a repair of $x$ if $x$ is deemed adversarial? Note that
we assume the availability of two models $f_1$ and $f_2$. In practice, multiple
models can be easily obtained by training with slightly different architectures,
or different training sets, or through model
mutation~\cite{wang2019adversarial}.

Figure~\ref{fig:framework} shows the overall workflow of our approach. Given an
input text $x$ and two models $(f_1, f_2)$, we first check whether $x$ is likely
adversarial. If the answer is positive, we apply adversarial perturbation to generate a
set of texts $X^*$ such that each $x \in X^*$ is slightly different from $x$ and
likely semantically equivalent to $x$. Afterwards, we apply a statistical
testing method to identify the most-likely correct label of $x$ (with a
guaranteed level of confidence) based on $X^*$ and output a text in $X^*$ which
is slightly different from $x$ as the repair. In the following paragraphs, we
present the details of each step.

\subsection{Adversarial Text Detection}
\label{sec:df}
Given an input text $x$, we first check whether it is adversarial (i.e., crafted by an attacker through adversarial perturbation). There are multiple methods for detecting adversarial perturbations in the image domain~\cite{zheng2018robust,wang2019adversarial}. The topic is relatively less studied in the text domain~\cite{xu2019adversarial}. Other than detection using a spell-checker, to the best of our knowledge, the only approach is the one mentioned in~\cite{rosenberg2019defense}, which focuses on re-training for improving robustness rather than detecting adversarial texts. In our work, we propose a detection method which is inspired by differential testing~\cite{mckeeman1998differential}.

Applying differential testing naively in our context (i.e., claim that $x$ is
adversarial if the labels generated by $f_1$ and $f_2$ are different) is
problematic. Adversarial samples in the image domain are known to have
transferability between different models~\cite{papernot2016transferability},
i.e., $f_1$ and $f_2$ may generate the same wrong label given the same
adversarial text. To examine how effective naive differential testing is, we
conduct an empirical study to evaluate the transferability of adversarial texts.
We train two different models, one TextCNN and one LSTM, for sentiment analysis
on three widely used standard datasets (i.e., NA~\cite{newsAggregator}, RTMR~\cite{rtmr} and
IMDB~\cite{imdb}). The performance of the models used for the clean text is shown
in Table~\ref{tab:model_acc}. Afterwards, we generate 1000 adversarial texts
on each dataset using TEXTBUGGER for the TextCNN model (respectively the LSTM
model) and check the accuracy of the LSTM model (respectively the TextCNN model)
with regards to these adversarial texts. Table~\ref{tab:trans} shows the results
where the second column shows the transferability of the adversarial texts
generated by attacking the TextCNN model and the third column shows that of the
adversarial texts generated by attacking the LSTM model. We confirm that
adversarial texts indeed transfer between different models (which is similar to
adversarial images~\cite{advtrans}). For instance, more than $40\%$ of
adversarial texts fool both models in the case of the NA and RTMR datasets.

\begin{table}[t]
  \centering
  \caption{Transferability rate  of adversarial texts.}
    {\small \begin{tabular}{|c|c|c|c|}
    \hline
    Dataset & TextCNN $\rightarrow$ LSTM & LSTM $\rightarrow$ TextCNN \\
    \hline
    NA      & 54\%      & 40.6\% \\ \hline
    RTMR    & 54.6\%  & 50.2\% \\ \hline
    IMDB    & 31.3\%  & 20.3\% \\
\hline
    \end{tabular}}\label{tab:trans}\end{table}

\begin{table}[t]
      \centering
      \caption{Performance of models used in our experiments.}
      \label{tab:model_acc}
      {\small \begin{tabular}{|c|c|c|c|}
      \hline
      Dataset               & Model  & Training Accuracy & Test Accuracy \\ \hline
      \multirow{2}{*}{NA} & TextCNN  & 98.20\%       & 89.21\% \\
                                                 & LSTM & 93.29\%      & 87.04\%\\ \hline
      \multirow{2}{*}{RTMR} & TextCNN  & 99.84\%       & 79.71\%    \\
                                                 & LSTM & 84.87\%      & 77.88\% \\ \hline
      \multirow{2}{*}{IMDB} & TextCNN  & 100\%       & 88.24\%     \\
                                                 & LSTM & 88.17\%      & 87.24\%     \\ \hline
      \end{tabular}}
\end{table}

We thus need a more reliable way to check whether $x$ is adversarial. Our remedy is to further measure the difference between the prediction distributions of the two models. Concretely, the output of a neural network for multi-class classification is a probability vector $f(x)=[p_0,p_1,\cdots,p_K]$, where $f$ is a model and $p_i$ is the probability of the input being class $i$ and $K$ is the total number of classes. We enhance differential testing by comparing the difference of two models' probability vectors. That is, the input $x$ is regarded adversarial if the difference of the probability vectors is larger than a threshold.

We adopt KL divergence ($D_{KL}$)~\cite{joyce2011kullback} to measure the difference between the two probability vectors. Formally, let $f_1(x) = [p_0,p_1,\cdots,p_K]$ and $f_2(x) = [q_0,q_1,\cdots,q_K]$.
\begin{align}
D_{KL}(f_1(x), f_2(x))=-\sum_{i=1}^K p_i \ln \frac{q_i }{p_i }
\end{align}
Hereafter, we write $D_{KL}(x)$ to denote $D_{KL}(f_1(x), f_2(x))$. Intuitively, $D_{KL}(x)$ is smaller if two distributions are more similar. Our hypothesis is that if the input is not adversarial, the probability vectors $f_1(x)$ and $f_2(x)$ should be similar and thus the difference $D_{KL}(x)$ should be small; otherwise it should be large. This is confirmed empirically as we show in Section~\ref{sec:exp}. 

Algorithm~\ref{alg:df-kl} shows the details of our adversarial sample detection algorithm, where $\epsilon$ is a threshold. An input $x$ is considered to be normal (refer to line 3) only if the labels generated by the two models are the same, and the $D_{KL}(x)$ is below the threshold. Otherwise, the input is regarded as adversarial. The remaining question is how to set the value of $\epsilon$, which we solve using the standard method of golden-section search as we discuss in Section~\ref{sec:exp}.

\begin{algorithm}[t]
\caption{$isAdversarial(x, f_1, f_2, \epsilon)$}
\label{alg:df-kl}
let $c_1$ be the output label according to $f_1(x) $\;
let $c_2$ be the output label according to $f_2(x) $\;
\If{$c_1 \equiv c_2$ and $ D_{KL}(x) < \epsilon$}{
  \Return{false}\;
}
\Return{true}\;
\end{algorithm}

\begin{example}
Table~\ref{tab:exmp1} shows an example on how our adversarial sample detection
algorithm works. The first row is a normal text from the RTMR dataset, and the
second row is an adversarial text generated using SEAs. The third and fourth
rows are the probability vectors generated by the two models respectively. The
task is sentiment analysis and thus there are two possible labels. Note that
while the original text is correctly labeled `positive', both models label the
adversarial text `negative'. The fifth row is the KL divergence of the two
probability vectors. Although the adversarial text fools both models, its
$D_{KL}(x)$ is larger than the threshold and thus is identified as adversarial.
Note that the threshold as shown in sixth row is selected empirically as we
explain in Section~\ref{sec:exp}.
\end{example}

\begin{table}[t]
\centering
\caption{KL divergence based adversarial sample detection.}
{\small \begin{tabular}{|l|l|}
\hline
\multirow{2}[1]{*}{Original text} &\textit{an appealingly juvenile trifle that}  \\
& \textit{delivers its share of laughs and smiles} \\ \hline
\multirow{2}[1]{*}{Adversarial text $x$} & \textit{a delightful childish trifle that can} \\
& \textit{bring laughter and a smile} \\ \hline
$f_1(x)$ TextCNN & $[0.9656, 0.03438]$  \\ \hline
$f_2(x)$ LSTM & $[0.68090, 0.3191]$ \\ \hline
$D_{KL}(x)$  & $0.2608$  \\ \hline
$\epsilon$ & $ 0.1110$  \\ \hline
\end{tabular}}
\label{tab:exmp1}
\end{table}

\subsection{Semantic-Preserving Perturbation}
\label{sec:spp}

Once we identify an adversarial text $x$, the next challenge is how to
automatically repair the input. In general, a repaired text $x'$ should satisfy
the following conditions: 1) $x'$ should be syntactically similar to $x$ and
semantically equivalent to $x$; 2) $x'$ should be classified as normal by our
adversarial sample detection algorithm; and 3) $x'$ should be labeled
correctly. In the following paragraphs, we describe how to systematically
generate a set of candidate repairs $X^*$ satisfying 1) and discuss how to
identify a repair among the candidates that satisfies all the conditions in
Section~\ref{sec:voting}.

We generate candidate repairs through perturbation, i.e., the same technique for
generating adversarial texts except that they are used in a positive way this
time. In particular, three different adversarial perturbation methods are
applied to generate syntactically similar and semantically equivalent texts.
Applying multiple perturbation methods allows us to compare their performance as
well as identify the right method for different usage scenarios. 

The first one is \emph{random perturbation}. Let $x = [w_1,w_2,\cdots,w_n]$
where $w_i$ is a word in the text $x$. To apply random perturbation on $x$, we
randomly select $g$ words in $x$ and replace them with their synonyms. Note that
to preserve the semantics, $g$ is typically small. In particular, for each
selected word $w_i$, we identify a ranked list $[w_{i1},w_{i2},\cdots,w_{iL}]$
of its synonyms of size $L$ according to their distances to $w_i$ measured in
the embedding space. As a result, we obtain $g^L$ perturbations. We refer
this method as \emph{RP} in the following paragraphs.

The second one is based on the idea of TEXTBUGGER with the \emph{Sub-W}
operation~\cite{textbugger}. That is, we first identify the important sentences,
and replace the important words in the sentence with their synonyms. Note that
different from TEXTBUGGER~\cite{textbugger}, our goal is to decrease $D_{KL}(x)$
so that the perturbed text passes the enhanced differential testing. Thus, we
evaluate the importance of a sentence and a word based on its effect on
$D_{KL}(x)$ (instead of the effect on the model prediction as
in~\cite{textbugger}). Concretely, to obtain the importance of a sentence $s_i$,
we calculate $D_{KL}(f_1(s_i), f_2(s_i))$. A sentence with a larger $D_{KL}$ is
considered more important. Within a sentence, we obtain the importance of a word
$w_j$ by measuring the $D_{KL}$ of the sentence with and without $w_j$, i.e.,
\begin{align}
\label{eq:klw}
D_{KL}(f_1(s_i), f_2(s_i))-D_{KL}(f_1(s_i\setminus w_j), f_2(s_i\setminus w_j))
\end{align}
A word causing a larger decrease of $D_{KL}$ is more important. Afterwards, the
important words are replaced with their synonyms to generate perturbations. The
details are shown in Algorithm~\ref{alg:tbp}. We refer this method to
\emph{SubW} in the following paragraphs.

\begin{algorithm}[t]
\caption{$TBPerturb(x, g, f_1, f_2)$}
\label{alg:tbp}
Let $C_s$ be the importance scores for each sentence in $x$\;
\For{$s_i\in x$}{
	$C_s(i) = D_{KL}(s_i)$\;
}
$S\leftarrow$ sort the sentences in $x$ according to $C_s$\;
\For{$s_i\in S_{ordered}$}{
	Let $C_w$ be the importance scores for each word in $s_i$\;
	\For{$w_j\in s_i$}{
		Compute $C_w(j)$ according to Eq.~\ref{eq:klw}\;
	}
	$W \leftarrow$ sort the words in $s_i$ according to $C_w$\;
}
\textit{combs} $\leftarrow$ select $g$ words according to $S$ and $W$\;
$x'\leftarrow$ replace each word $w\in combs$ in $x$ with synonyms\;
\Return{$x'$}\;
\end{algorithm}

The third one is to generate semantic-preserving texts using NMT in a way
similar to SEAs. Formally, an NMT is a function $T(s,d,x):X_s\to X_d$, where $s$
is the source language, $d$ is the destination language and $x$ is the input
text. The basic idea is to translate the input text into another language and
then translate it back, i.e., the new text is $x' = T(d, s, T(s, d, x))$. By
varying the target language $d$ (e.g., French and Germany), we can generate
multiple perturbations this way. Furthermore, it is possible to translate across
multiple languages to generate even more perturbations. For instance, with two
target languages $d_1$ and $d_2$, we can generate $x' = T(d_2, s, T(d_1, d_2,
T(s, d_1, x)))$ as perturbations. Note that compared to perturbations generated
using random perturbation or Algorithm~\ref{alg:tbp}, the texts generated
through NMT might have a different length or syntactical structures, which results
in a larger distance in the embedding space. It would be interesting to evaluate
whether such a difference affects the effectiveness of our approach. For the
sake of convenience, we refer this paraphrase-based perturbation approach
to \emph{ParaPer} in the following paragraphs.

\begin{table*}[t]
\centering
\footnotesize
\caption{Example of semantic-preserving perturbation}
\begin{tabular}{|l|l|}
\hline
Original text &  \textit{a delightful childish trifle that can bring laughter and a smile}  \\ \hline
Perturbed text with \emph{RP} & \textit{a \textbf{charming silly trifles} that can bring laughter and \textbf{another} smile
}\\ \hline
Perturbed text with \emph{SubW} & \textit{a delightful \textbf{childlike trifling} that can bring laughter and a smile
}\\ \hline
Source language & English \\
Target language &  Hungarian \\
Translated text & \textit{Egy elragadó gyerekes apróság, ami nevetést és mosolyt hoz}  \\
Perturbed text with \emph{ParaPer} & \textit{A delightful childish \textbf{little thing} that can bring laughter and \textbf{smiles}}\\ \hline
\end{tabular}
\label{tab:exmp2}
\end{table*}

\begin{example}
Table~\ref{tab:exmp2} shows an example of our semantic-preserving perturbation with different methods. The original text is shown in the first row (i.e., the adversarial text shown in Table~\ref{tab:exmp1}). The perturbed texts using random perturbation and Algorithm~\ref{alg:tbp} are shown in the second and third row. For SEAs perturbation, we translate the original text into Hungarian, which is then translated back into English (shown in the last row). Note that the perturbed text generated by SEAs may have a different number of words.
\end{example}

 \subsection{Voting for the correct label} \label{sec:voting} After generating a
set of texts $X^*$ which are slightly mutated from $x$ and yet are semantically
equivalent to $x$, our next step is to identify a member $x'$ of $X^*$ that
satisfies 2) $x'$ should be classified as normal by our adversarial sample
detection algorithm and 3) $x'$ is correctly labeled. Satisfying 2) is
straightforward. That is, we filter those in $X^*$ which are determined to be
adversarial using Algorithm~\ref{alg:df-kl}. The result is a set $X^*$ such that
every $y$ in $X^*$ satisfies $f_1(y) = f_2(y)$ and $D_{KL}(y) < \epsilon$.
Satisfying 3) requires us to know what the correct label is. Our idea is that we can `vote' and decide
on the correct label. Our hypothesis is that the majority of texts in $X^*$ are
likely classified correctly and thus a democratic decision would be correct.
This idea is inspired by the observation made in~\cite{wang2019adversarial},
which shows that adversarial samples (with wrong labels) in the image domain
have a high label-change rate when perturbations are
applied~\cite{wang2019adversarial}. In other words, perturbing adversarial
samples would often restore the correct label. One interpretation is that
adversarial samples are generated by perturbing normal samples just enough to
cross the classification boundary, and thus a slight mutation often restores the
original label. We evaluate this hypothesis empirically in
Section~\ref{sec:exp}.

Based on the hypothesis, we formulate the problem as a statistical testing problem. That is, we present it with a set of hypotheses (e.g., the correct label of a text is $c_i$ where $c_i$ is one of the labels) and the problem is to identify the hypothesis which is most likely true with statistical confidence. To solve the problem, we adopt hypothesis testing~\cite{shaffer1995multiple} to guarantee that the probability of choosing the correct label is beyond a threshold, say $\rho$. That is, given a label $c_i$, we systematically test the null hypothesis ($H_0$) and the alternative hypothesis ($H_1$) which are defined as follows.
\begin{align}
\label{eq:ht1}
H_0(c_i): P(f(x) = c_i) \geq \rho \\
\label{eq:ht2}
H_1(c_i): P(f(x) = c_i) < \rho
\end{align}
, where $P(f(x) = c_i)$ is the probability that the true label of $x$ is $c_i$. Given $X^*$ which contains only texts that are semantically equivalent to $x$, we estimate $P(f(x) = c_i)$ as follows.
\begin{align}
\label{hprb}
P(f(x) = c_i) = \frac{|y \in X^* \land f_1(y) = c_i |}{|X^*|}
\end{align}
Note that all texts in $X^*$ have the same label according to model $f_1$ and $f_2$ after filtering as mentioned above. We remark as long as we set $\rho$ to be more than $0.5$, we guarantee that only one $H_0(c_i)$ for some $c_i$ is accepted.
In general, given a limited number of perturbations, it might be possible that none of the $H_0(c_i)$ is accepted.

Since there are multiple labels, we maintain a pair of hypotheses for each $c_i
\in C$ and perform a hypothesis testing procedure for every pair. There are two
ways for performing hypothesis testing. One is the fixed-size sampling test
(FSST), which performs the test on a fixed number of samples. That is, we first
generate a set $X^*$ with a sufficiently large number of samples, calculate
$P(f_1(x) = c_i)$ for each label $c_i$ according to (4), and then compare the
result with $\rho$. The drawback of FSST is that we must determine what is the
minimum number of samples required such that the error bounds are satisfied.
Typically, FSST requires a large number of samples~\cite{fsst}.

In general, the more samples that we use, the more accurate the result would be.
On the other hand, the more samples required, the more computational overhead
there is, which may be problematic if such repairing is to be carried out in an
online manner (e.g., for suggesting repaired forum posts timely). We thus propose to
use the sequential probability ratio test (SPRT~\cite{sprt}), which dynamically
determines the number of samples required and is known to be faster than
FSST~\cite{wald1973sequential}. Central to SPRT is to repeatedly sample 
until enough evidence is accumulated to make a decision (accepting either hypothesis).

Algorithm~\ref{alg:ht} shows the details on how SPRT is applied in our work to
decide whether to accept hypothesis $H_0(c_i)$ or not for label $c_i$. Note that
$\alpha$ is the probability of the case in which $H_0$ is reject while $H_0$ is
true (a.k.a. Type $\rom{1}$ error), $\beta$ is the probability of the case in
which $H_1$ is reject while $H_1$ is true (a.k.a. Type $\rom{2}$ error). $\rho$ is
the confidence threshold described before and $\sigma$ is the indifference
interval used to relax the threshold. We then test hypotheses $H_0(c_i): P(f(x)= c_i) \geq p_0$ and $H_1(c_i): P(f(x) = c_i) < p_1$ where $p_0 = \rho + \sigma$ 
and $p_1 = \rho - \sigma$. At line 6, we compute the likelihood ratio of SPRT
which is defined as follows~\cite{wald1973sequential}.
\begin{align}
\label{eq:sprtr}
Pr(z,k, p_0,p_1) = \frac{p_1^{z}(1-p_1)^{k-z}}{p_0^{z} (1-p_0)^{k-z}}
 \end{align}
At line 7, we check whether the ratio is no larger than
$\frac{\beta}{1-\alpha}$. If it is the case, the hypothesis $H_0(c_i) \geq p_0$
is accepted and report the label $c_i$ as the true label with error bounded by
$\beta$. If the ratio is no less than $\frac{1-\beta}{\alpha}$, we then accept
$H_1(c_i) \leq p_1$ at line 11 and report the label $c_i$ is not the true label with error bounded by
$\alpha$. Otherwise, it is inconclusive (i.e., more samples are required).

\begin{algorithm}[t]
\caption{$hypTest(c_i, X^*, f_1,\alpha,\beta,\sigma,\rho)$}
\label{alg:ht}
Let $k$ be the size of $X^*$\;
Let $z$ be the size of $\{y | y \in X^* \land f_1(y)=c\}$\;
Let $\alpha,\beta,\sigma,\rho$ be the parameter of hypothesis testing\;
 $p_0 = \rho + \sigma $\;
 $p_1 = \rho - \sigma$\;
$sprt\_ratio \leftarrow Pr(z,k,p_0,p_1)$\;
 \If{$sprt\_ratio \leq \frac{\beta}{1-\alpha }$}{
Accept the hypothesis that $H(c) \geq p_0$ \;
\Return\; 
}
\If{$sprt\_ratio \geq \frac{1 - \beta}{\alpha }$}{
  Accept the hypothesis that $H(c) \leq p_1$ \;
\Return \;
}
  \Return \textit{Inconclusive};	
\end{algorithm}

\begin{algorithm}[t]
\caption{$Repair(x, f_1, f_2, \epsilon, \alpha, \beta, \sigma, \rho)$}
\label{alg:repair}
\If{$isAdversarial(x,f_1,f_2, \epsilon)$}{
Let $X^*$ be an empty set\;
    Let $C=\emptyset$ be a set of possible labels\;
    Let $D=\emptyset$ be a set of rejected labels\;
\If{$f_1(x) = f_2(x)$}{
       $D = D \cup \{f_1(x)\}$\;
    }
    \While{true}{
Let $y$ = $perturb(x)$\;
        \If{$isAdversarial(y, f_1, f_2, \epsilon)$} { continue\; }
$c=f_1(y)$\;
             $X^*= X^*\cup\{y\}$\;
\If{$c\notin C$ and $c \notin D$}{
                $C=C\cup\{c\}$\;
}
\For {each $c_i$ in $C$ and $c_i \notin D$} {
                Let $co$ be $hypTest(c_i, X^*, f_1, \alpha, \beta, \sigma, \rho)$\;
                \If{$co$ is accept}{
                    \Return $x' \in X^*$ s.t.~$f_1(x')=c_i$\;
                }
                \If{$co$ is reject}{
                    $D=D\cup\{c_i\}$\;
                }
            }
}
        }
\Return $x$\;
\end{algorithm}

\subsection{Overall Algorithm}
The overall algorithm is shown in Alg.~\ref{alg:repair}. The inputs include an
input text $x$, a pair of NNs $f_1$ and $f_2$, a threshold $\epsilon$, the
parameters required for hypothesis testing, and a threshold $\rho$. We first
check whether $x$ is adversarial or not at line 1 using
Algorithm~\ref{alg:df-kl}. If it is a normal text, $x$ is returned without any
modification. If it is adversarial and the labels from the two models are the
same, the label is added into $D$ (i.e., a set of labels which we know are
incorrect) at line 6 since we know that it is not the correct label.

The loop from line 7 to 20 then aims to repair $x$. We first obtain a
semantic-preserving perturbation of $x$ at line 8. Note that function
$perturb(x)$ can be implemented using either RP,
SubW or ParaPer as we discussed in Section~\ref{sec:spp}. We
then check whether the newly generated text $y$ is adversarial. If it is, we
generate another one until a perturbed text $y$ which is determined to be normal
is generated. If $y$ has a label which is never seen before, we add the label to
$C$ which is a set of potentially correct labels for $x$. Afterwards, for each
potential label $c_i$ in $C$, we conduct hypothesis testing using
Algorithm~\ref{alg:ht} at line 16. If the conclusion is to accept $H_0(c_i)$, we
identify a text in $X^*$ which has the label $c_i$ as a repair of $x$ and return
it. If the conclusion is to reject $H_0(c_i)$, the label $c_i$ is added into $D$
(so that it is never tested again) and we continue with the next iteration.
Otherwise, if it is inclusive, we continue with the next iteration. Note that to
reduce the computational overhead, we conduct hypothesis testing in a
\emph{lazy} way. That is, we maintain a set of witnessed labels $C$ (which is
initially empty) and only test those in $C$. Furthermore, we maintain a set of
rejected labels $D$ so that as soon as a label is rejected, it is never tested
again.

Algorithm~\ref{alg:repair} always terminates. Given any label $c_i$,
Algorithm~\ref{alg:ht} always terminate since SPRT is guaranteed to terminate
with probability 1~\cite{wald1973sequential}. As there are finitely many labels,
and each label is tested by Algorithm~\ref{alg:ht} once, it follows
Algorithm~\ref{alg:repair} always terminates.

\begin{figure}[t]
\centering
\includegraphics[width=0.48\textwidth]{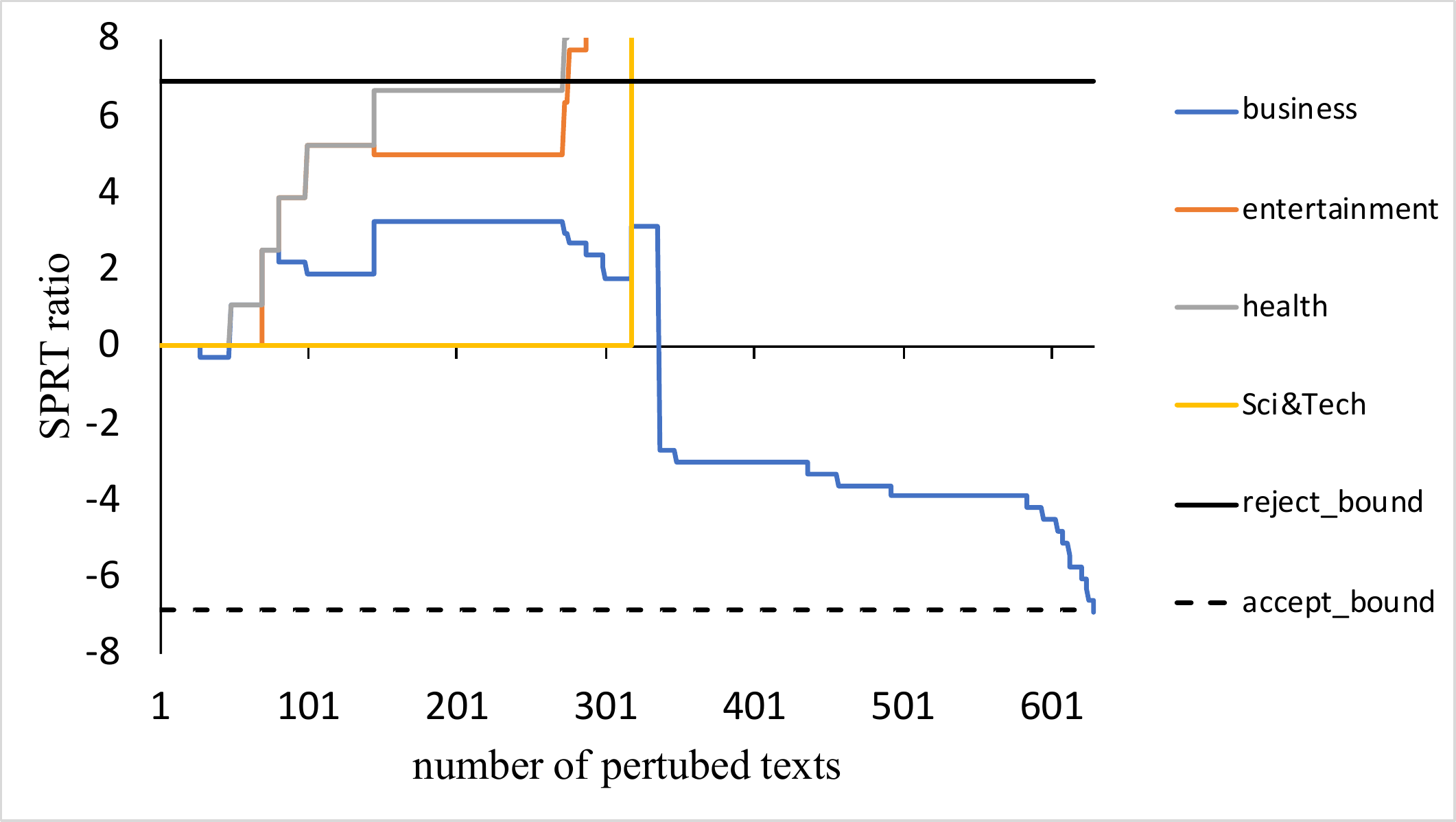}
\caption{Example of hypothesis testing}
\label{fig:exmp3}
\end{figure}

\begin{example}
We show how our algorithm works using an example. We are given two models, i.e.,
a TextCNN ($f_1$) and an LSTM ($f_2$), for topic labeling trained on the News
Aggregator Dataset~\cite{newsAggregator}. Given the original text \textit{``US
city moves to stop Monkey Parking''} in the dataset, an adversarial text
\textit{``States city turns to stop Monkey Parking''} is generated using
TEXTBUGGER. Note that the original correct label is ``business'' while the
adversarial text is classified as ``Sci\&Tec''. We feed the adversarial text
into Algorithm~\ref{alg:repair} and adopt the SEAs perturbation method to
generate perturbed texts. The parameters $\alpha, \beta, \sigma$ and $\rho$ in
Algorithm~\ref{alg:ht} are $0.001, 0.001, 0.15, 0.8$ respectively. The
acceptance bound and rejection bound are consequently $-6.9068$ and $6.9068$
respectively. Our approach works as follows. The text is identified to be
adversarial at line 1 in Algorithm~\ref{alg:repair}. Then, SEAs is applied to
generate perturbed texts for hypothesis testing to vote for the correct label.
At the \textit{31-th} attempts, the algorithm starts a hypothesis testing
procedure for label ``business''. The testing procedures for label ``health" and
``entertainment" are started at \textit{51-th} and \textit{73-th} attempts
respectively. Label ``health" and ``entertainment" are rejected at the
\textit{275-th} and \textit{278-th} attempts when the SPRT ratio are 8.0155 and
7.7132 respectively. The algorithm immediately rejected ``Sci\&Tec'' when the
label first appeared since there are already many perturbed texts with label
``business". Label ``business" is finally accepted at the \textit{627-th}
attempt when its SPRT ratio is -6.9525. Figure~\ref{fig:exmp3} shows how the
confidence of each label changes with an increasing number of perturbed texts.
\end{example}
  \section{Experiments}
\label{sec:exp}
We have implemented our approach as a prototype targeting LSTM and TextCNN models trained for NLP classification tasks. The implementation is in PyTorch~\footnote{https://pytorch.org/} with about 5000 lines of code. In the following paragraphs, we conduct multiple experiments to answer the following research questions (RQ).
\begin{itemize}
    \item RQ1: Is KL divergence useful in detecting adversarial texts?
    \item RQ2: Is our hypothesis for voting justified?
    \item RQ3: Is our approach effective at fixing adversarial texts?
    \item RQ4: What is the time overhead of our approach?
\end{itemize}
RQ1 is important as detecting adversarial texts is a prerequisite for our approach. Only with an effective adversarial sample detection approach, our repairing procedure can be triggered effectively. RQ2 asks whether our hypothesis that most of the texts generated through perturbing an adversarial text are normal is valid or not. Note that this would justify our approach for repairing. RQ3 then checks whether the overall approach would effectively repair adversarial texts. Lastly, we evaluate the time efficiency of our approach in order to see whether it is applicable in a time-constrained setting like online repairing. All experiments are carried out on a workstation with 1 Intel Xeon 3.50GHz CPU, 64GB system memory and 1 NVIDIA GTX 1080Ti GPU.
%The tool together with all the evaluation details are released onlines~\footnote{\url{https://github.com/dgl-prc/adv_text_repair}}.

\subsection{Experiment Settings}
\label{sec:es}
We conduct our experiments on the following three popular real-world datasets which include the two used by TEXTBUGGER~\cite{textbugger}.
\begin{itemize}
    \item \emph{News Aggregator (NA) Dataset}~\cite{newsAggregator} This dataset contains 422419 news stories in four categories: business, science and technology, entertainment, and health. For the sake of efficiency, we randomly take $10\%$ of the dataset for our experiment. The task is multi-topic labeling.
    \item \emph{Rotten Tomatoes Movie Review} This dataset is another movie review dataset collected from Rotten Tomatoes pages~\cite{rtmr} for sentiment analysis, which contains 5331 positive and 5331 negative sentences.
    \item \emph{IMDB} This dataset is a widely used dataset for sentiment analysis classification and contains $50,000$ movie reviews from IMDB~\cite{imdb} which are equally split into a training set and a test set. In total, there are 25k positive reviews and 25k negative reviews. Following~\cite{textbugger}, we randomly select $20\%$ of the training data for training the NNs. In the following paragraphs, unless stated otherwise, we follow the standard splitting to have 80\% of the dataset for training and 20\% for testing.
\end{itemize}

We adopt two heterogeneous NNs widely used for text classification as the target
models: LSTM~\cite{lstm} and TextCNN~\cite{textcnn}. LSTM is a classical
recurrent neural network model used to deal with sequential data in natural
language processing. In our case, LSTM is a vanilla one as used
in~\cite{zhang2015character}. TextCNN is a convolutional neural network for text
classification. TextCNN has four different types based on the strategy of using
word vectors: \textit{CNN-rand}, \textit{CNN-static}, \textit{CNN-non-static}
and \textit{CNN-multichannel}. We choose \textit{CNN-static} since we do not
need to modify the pre-trained word vectors. We follow the configuration of
TextCNN in~\cite{textcnn}. To train both models for each of the three datasets,
we first transform each word into a 300-dimensions numerical vector using the
pre-trained word vectors GloVe~\cite{glove}. The performance of our trained
models is presented in Table~\ref{tab:model_acc}, which is comparable to the
state-of-the-art.

We adopt two state-of-the-art approaches to generate adversarial texts, i.e.,
TEXTBUGGER with \textit{Sub-W} and SEAs. For each model, we randomly select 300
texts from the dataset and apply both attacks to generate adversarial texts. The
3rd column and 4th column in Table~\ref{tab:advtexts} summarize the number of
adversarial texts generated using each method. Note that the number is smaller
than 300 since the attack is not always successful. In total, we have
3642 adversarial texts generated using two different attacking methods on
six models. 
% \gl{and for IMDB the adversarial
% texts are generated from benign texts which have a maximum length of 100
% considering that the costs of using industrial API}.

\begin{figure}[t]
  \centering
  \includegraphics[width=0.4\textwidth]{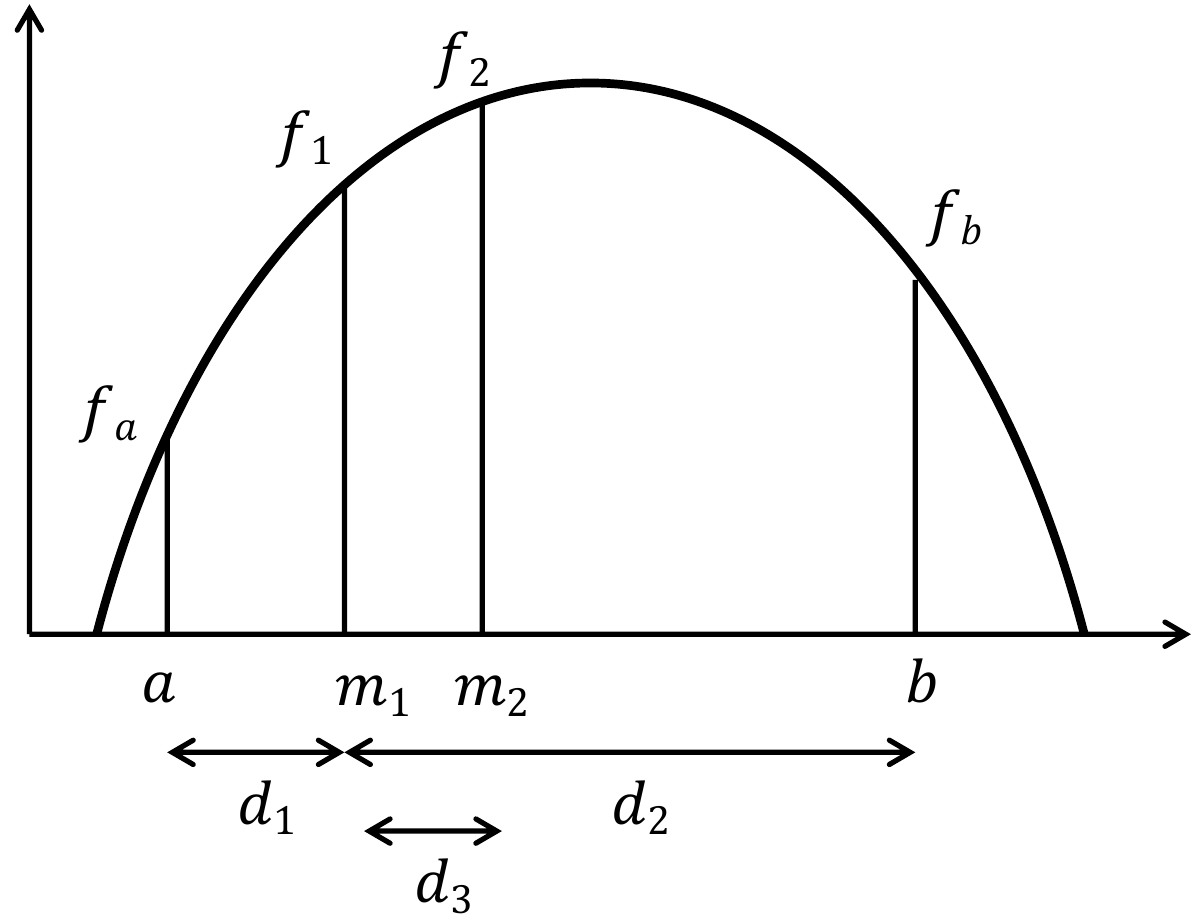}
  \caption{Example of a golden section search. The initial interval is $[a, b]$
  and the first split point is $m_1$. In the interval $[m_1, b]$, we choose the
  second split point $m_2$. Since the $f_2>f_1$, the triplet $(m_1,m_2, b)$ is chosen for the next iteration. Note that $d_2:d_1=(d_2-d_3):d_3$ is the golden ration.}
  \label{fig:golden}
\end{figure}

\begin{table}[t]
\centering
\caption{The number of generated adversarial texts and Threshold ($\epsilon$) in
Algorithm~\ref{alg:df-kl}, where TextB is short for TEXTBUGGER and TextF is short for TEXTFOOLER. Without mentioning in the follow-up table, the same short form denotes the same thing.}
\label{tab:advtexts}
\begin{tabular}{|c|c|c|c|c|c|}
\hline
\multirow{2}{*}{Dataset} & \multirow{2}{*}{Model}           & \multicolumn{3}{c|}{Attack}    &  \multirow{2}{*}{Threshold ($\epsilon$) }   \\ \cline{3-5}
                         				  &                      & SEAs   & TextB  & TextF 	 & 			                      \\ \hline
\multicolumn{1}{|c|}{\multirow{2}{*}{NA}}   & TextCNN    & 149    &165     &  99         & 0.0288	             			                   \\
\multicolumn{1}{|c|}{}                      & LSTM       & 159    &182     &  97         & 0.0266					          \\
\multicolumn{1}{|c|}{\multirow{2}{*}{RTMR}} & TextCNN    &222     &229     &  273         & 0.0655 \\
\multicolumn{1}{|c|}{}                      & LSTM       &210     &231     &  279        &  0.111	\\
\multicolumn{1}{|c|}{\multirow{2}{*}{IMDB}} & TextCNN    &168     &227     &  229        & 0.1593   \\
\multicolumn{1}{|c|}{}                      & LSTM       &166     &267     &  280       &  0.1806				            \\  \hline
\end{tabular}
\end{table}

To generate perturbations using random perturbations and
Algorithm~\ref{alg:tbp}, we limit the maximum number of words to be replaced to
be 4 so that the resultant text is likely semantic-preserving. To obtain the
synonyms of a chosen word, we use
\textit{gensim}~\footnote{https://radimrehurek.com/gensim/}, which is an
open-source library to find the most similar words in the word embedding space.
To perform SEAs perturbation, we utilize the NMTs from an online Translation API
service~\footnote{http://api.fanyi.baidu.com/api/trans/product/index}.

\subsection{Research Questions}

\begin{table}[t]
  \centering
  \caption{Effectiveness of adversarial detection over different adversarial texts. ``BL" denotes the baseline detection method, i.e., vanilla differential 
         testing, while ``KL-D" denotes our KL-divergence based approach. ``dr" and ``fp" denotes the detection rate (\%) and false positive rate (\%). ``TextB" and ``TextF'' is short for TEXTBUGGER and TEXTFOOLER respectively.}
        %  Without mentioning in the follow-up table, the same short form denotes the same thing.
   \begin{tabular}{|c|c|c|c|c|c|c|c|c|c|}
    \hline
    \multirow{3}[6]{*}{Attack} & \multirow{3}[6]{*}{Dataset} & \multicolumn{4}{c|}{TextCNN}  & \multicolumn{4}{c|}{LSTM} \\ \cline{3-10}
     &       & \multicolumn{2}{c|}{BL} & \multicolumn{2}{c|}{KL-D} & \multicolumn{2}{c|}{BL} & \multicolumn{2}{c|}{KL-D} \\ \cline{3-10}
    &       & dr   & \multicolumn{1}{l|}{fp} & dr  & \multicolumn{1}{l|}{fp} & dr   & \multicolumn{1}{l|}{fp} & dr   & \multicolumn{1}{l|}{fp} \\\hline

    \multirow{4}{*}{SEAs} & {NA}   & {47} & {10} & {88} & {19} & {59} & {2} & {92} & {17} \\

                          & {RTMR} & {44} & {19} & {68} & {36} & {44} & {18} & {63} & {29} \\

                          & {IMDB} & {64} & {14} & {76} & {25} & {59} & {7} & {73} & {18} \\ \cline{2-10} 
                          & {Avg} & {\textbf{52}} & {\textbf{14}} & {\textbf{77}} & {\textbf{27}} & {\textbf{54}} & {\textbf{9}} & {\textbf{76}} & {\textbf{21}} \\ \hline

    \multirow{4}{*}{TextB} & {NA} & {68} &  {7}  &{93}   & {19} & {73} & {1} & {93} & {18} \\

                          & {RTMR} & {54} & {16} &{76}  & {33} & {68} & {12} & {79} & {25} \\

                          & {IMDB} & {84} & {12} &{89}  & {23} & {85} & {7} & {93} & {17} \\ \cline{2-10}

                          & {Avg} & {\textbf{69}} & {\textbf{12}} & {\textbf{86}} & {\textbf{25}} & {\textbf{75}} & {\textbf{7}} & {\textbf{88}} & {\textbf{20}} \\ \hline

   {\multirow{4}{*}{TextF}} & NA     &62  &9      &95    &23     &65  &6  &95   &17  \\

                            & {RTMR} &41  &16     &67    &39     &50  &13  &68  &28 \\
 
                            & {IMDB} &77  &9      &89    &19      &79  &9  &90  &19  \\ \cline{2-10}
 
                            & {Avg}  &\textbf{60} 	&\textbf{11} 	&\textbf{84} 	&\textbf{27} 	&\textbf{65} 	&\textbf{9}	&\textbf{84} 	&\textbf{21} \\
                            
 \hline
    % \multicolumn{2}{|c|}{Avg} & \textbf{58} & \textbf{13} & \textbf{82} & \textbf{27} & \textbf{64} & \textbf{8 }& \textbf{82} & \textbf{22} \\
    \multicolumn{2}{|c|}{Avg} &\textbf{60} 	&\textbf{12} 	&\textbf{82} 	&\textbf{26} 	&\textbf{65} 	&\textbf{8}	&\textbf{83} 	&\textbf{21} \\
    \hline
 \end{tabular}
\label{tab:addlabel}
\end{table}

\begin{table}[t]
  \centering
  \caption{Effectiveness of detection and repair when the adversarial texts are from a
  third model. The ``source model" in the first column refers to the model the
  adversarial texts generated for. The adversarial texts are generated from NA
  dataset with TEXTBUGGER.} 
  \label{tab:taa}
    \begin{tabular}{|c|c|c|c|c|}
    \hline
    \multirow{2}{*}{Source model} & \multirow{2}{*}{Models of detector}  & \multicolumn{2}{c|}{Detection results} & \multirow{2}{*}{\begin{tabular}[c]{@{}l@{}}Repair\\accuracy(\%) \end{tabular}} \\ \cline{3-4}
    &                             & dr (\%)  & fp (\%) & \\ \hline
    BiLSTM       & TextCNN, LSTM       & 89  &  21   & 60 \\ \hline
    LSTM         & TextCNN, BiLSTM     & 92  &  23   & 61.4  \\ \hline
    TextCNN      & LSTM, BiLSTM        & 87  &  25   & 46.1  \\ \hline
    \multicolumn{2}{|c|}{Avg}          & 89  & 23    & 55.83    \\ \hline
    \end{tabular}
\end{table}

\begin{table}[t]
  \centering
  \caption{Success rate (\%) of white-box attack for our detection approach. ``\#models" denotes the number of models to attack simultaneously.}
   \begin{tabular}{|l|l|l|l|l|}
    \hline
          \#models & Model(s) & NA & MR & IMDB \\
    \hline
    \multirow{4}[8]{*}{One} & TextCNN      & 57.6 & 74.3  & 66.30 \\
                            & LSTM         & 66.3 & 78.3  & 96.60 \\
                            & BiLSTM       & 69   &  79.78  & 97.1 \\ \cline{2-5}
                  & Avg   & \textbf{64.3}  & \textbf{77.46} & \textbf{86.16} \\
    \hline
    \multirow{4}[8]{*}{Two} & TextCNN+LSTM     & 1.6  & 28.36   & 6.5 \\
                            & TextCNN+BiLSTM   & 0.8  & 24.4    & 10.7 \\
                            & LSTM+BiLSTM      & 2.8  & 24.27   & 43.8 \\ \cline{2-5}
                & Avg   & \textbf{1.73}    & \textbf{25.68}  & \textbf{20.33} \\
    \hline
    Three& TextCNN+LSTM+BiLSTM   & \textbf{0.2}  & \textbf{12.4}    & \textbf{0} \\
    \hline
    \end{tabular}
  \label{tab:muli-dect}
\end{table}

  \begin{table}[t]
      \centering
      \caption{The impact of the architecture on our detection approach and repair approach. }
      \begin{tabular}{|c|c|c|c|c|}
      \hline
      \multirow{2}{*}{Source model} & \multirow{2}{*}{Models of detector} & \multicolumn{2}{c|}{Detection results} & \multirow{2}{*}{\begin{tabular}[c]{@{}l@{}}Repair\\accuracy(\%) \end{tabular}} \\ \cline{3-4}
                                    &                                     & dr (\%)    & fp  (\%)    &     \\ \hline
      BiLSTM                        & \multirow{2}{*}{BiLSTM,  FastText}  & 95     & 22   & 57.8                              \\ \cline{1-1} \cline{3-5} 
      FastText                      &                                     & 98     & 26   & 74    \\ \hline
      \end{tabular}
      \label{tab:ar_im}
  \end{table}

\emph{RQ1: Is KL divergence useful in detecting adversarial samples?} To answer
the question, we measure the accuracy of detecting adversarial texts using
Algorithm~\ref{alg:df-kl} and compare that to the alternative approach. Note
that to apply Algorithm~\ref{alg:df-kl}, we must first select the threshold
$\epsilon$. Ideally, the threshold $\epsilon$ should be chosen such that
$D_{KL}$ of normal texts are smaller than $\epsilon$ and $D_{KL}$ of adversarial
texts are larger than $\epsilon$ (in which case the accuracy of the detection is
1). 

In our implementation, we adopt golden-section
search~\cite{kiefer1953sequential,avriel1966optimally,wang2017should} which is
commonly used to find the extremum of a function (i.e., accuracy of adversarial
text detection) to identify $\epsilon$. The search procedure consists of four
steps: 1) given a search interval of $D_{KL}$, e.g, $[a, b]$, we first split the
interval $[a, b]$ according to the golden ratio to obtain an initial triplet
$(a, m_1, b)$, where $m_1$ is the split point; 2) next, we recursively split the
larger interval, i.e., $[m_1, b]$, according to the golden ratio and let $m_2$
be the splitting point; 3) after that, we calculate the accuracy with threshold
$m_1$, $m_2$, and $b$, respectively. If the accuracy of $m_2$ is lower than that
of $m_1$, we then take the triplet $(a, m_1, m_2)$ as the new triplet; otherwise
we take the triplet ($m_1$, $m_2$, $b$) as the new triplet; 4) once a new
triplet identified, we go to step 2) and repeat the above procedure until the
search interval is sufficiently small. Fig~\ref{fig:golden} exemplifies how the search algorithm works. In our experiments, we empirically set
initial range of $D_{KL}$ as [0, 10], and for each dataset and the target model,
we select all the wrongly classified samples and the identical number of
correctly classified samples to form the test set to obtain the detection
accuracy. We summarize our search results on each dataset in the last column of
Table~\ref{tab:advtexts}.

After setting $\epsilon$ as discussed above, we systematically apply
Algorithm~\ref{alg:df-kl} to a set of texts which mixes all 3642 adversarial
texts and an equal number of normal texts. For the baseline comparison, we compare
our algorithm with the alternative approach which simply checks whether the two
models agree on the output labels. The text is regarded as adversarial if the
answer is no. Otherwise, the text is regarded as normal. To the best of our
knowledge, there are no existing methods or tools which are available for
detecting adversarial texts. Note that the tool mentioned
in~\cite{rosenberg2019defense} is not available.

The results are summarized in Table~\ref{tab:addlabel} where column BL are the
results of the baseline approach and column KL-D is the results of our approach.
Furthermore, column `dr', i.e., the detection rate, denotes the percentage of
adversarial texts which are detected; and column `fp', i.e., the false positive
rate, denotes that out of all the text identified as adversarial, how many
percent are actually normal texts. Note that all numbers are percentile. It can
be observed that our approach detects most of the adversarial texts. Our
algorithm significantly outperforms the baseline for all datasets and models,
i.e., on average 76.5\% of the adversarial texts generated by SEAs are detected,
87\% of TEXTBUGGER and 84\% of TEXTFOOLER are detected, which are 23.5\%, 15\%
and 21.5\% higher than that of the baseline respectively. In particular, the
detection rate is $41\%$ higher for the TextCNN model with SEAs as the attacking
method on the NA dataset. This shows that our adversarial detection algorithm
effectively addresses the problem due to the transferability of adversarial
texts. 

We also observe that Algorithm~\ref{alg:df-kl} achieves a higher detection rate
in detecting adversarial texts generated by TEXTBUGGER than detecting those
generated by SEAs, i.e., $10.5\%$ higher on average. One possible explanation is
that the adversarial texts generated by TEXTBUGGER are likely to have a
relatively small `distance' from the original text. In comparison, adversarial
texts generated by SEAs may have different structures (after two translations)
and thus a relatively large distance to the original text. We also notice that the
detection rate of adversarial texts generated by TEXTBUGGER is close to that of
adversarial texts generated by TEXTFOOLER, i.e., only about a 3\% gap. This is
not surprising since the two methods in crafting adversarial text are pretty
similar as depicted in Section~\ref{sec:back}. Furthermore, since the
adversarial texts generated by TEXTFOOLER are more natural (it not only checks
the semantic similarity but also takes the part-of-speech into account when
replacing words.), these adversarial texts thus are more difficult to detect.

On average, our method has false positive rate of 26\% for the adversarial texts
generated by attacking the TextCNN model and 21\% for those generated by
attacking the LSTM model, which is higher than the baseline approach. Consider
that the baseline approach overlooks many adversarial texts (e.g., almost half
of those generated by SEAs), we believe this is acceptable. In addition, our
framework aims to automatically repair the ``alarms" and thus some false positives
can be eliminated by the subsequent repair. Later, we will show the
effectiveness of our approach on handling the false positive samples in RQ3.

\textit{Effectiveness on a Third Model.} In the above experiments, we assume
that the adversarial samples are from one of the two models used in detection. A
natural question is that if our approach can deal with the adversarial texts
from a model which is different from the two models used in detection. To answer
this question, we introduce a third model, i.e., BiLSTM~\cite{bilstm} which
consists of two LSTMs: one taking the input in a forward direction, and the
other in a backwards direction. Then, we apply our approach to detect the
adversarial texts generated from one model and use the other two for the
detection. For every third model, we take 1000 adversarial texts (generated by
TEXTBUGGER) and 1000 normal texts for the experiments. The results are
summarized in Table~\ref{tab:taa}. We can observe that the average detection
rate is 89\%, which suggests our approach can effectively identify the
adversarial texts from an unseen model.

% Then, we detect the adversarial texts generated
% for one model with our approach consisting of another two models. For each
% target model, i.e., a third model, we take 1000 adversarial texts (generated by
% TEXTBUGGER) and 1000 normal texts from NA dataset to conduct the experiments.

\textit{Effectiveness on Defending against White-box Attacks.} Another concern
is that if the attacker is aware of our detection method, then he or she may
devise an approach to generate adversarial texts evading the detection, i.e.,
generate adversarial texts which are claimed normal by
Algorithm~\ref{alg:df-kl}, which can be regarded as the white-box attack. To
answer this question, we conduct the following experiment. We first modify
TEXTBUGGER so that it aims to evade the detection by Algorithm~\ref{alg:df-kl},
i.e., by changing the importance score of sentences and words so as to generate
adversarial texts which keep lowering $D_{KL}$. Then, we apply it to generate
adversarial texts based on 1000 benign samples of each dataset and report the
success rate (i.e., how often it evades the detection). The results are shown in
Table~\ref{tab:muli-dect}. For comparison, the row `one' shows the success rates
of attack without detection and the row `two' shows that with detection using
two models. The last row `three' is the success rate of the attack if we adopt
three models for detection, i.e., a text is adversarial if the $D_{KL}$ of any
of the two models is more than the threshold. It can be observed that the
success rate drops significantly with detection using two models, and drops even
further if three models are used. In particular, for dataset NA and IMDB, almost none of the attack is
successful. We thus conclude that our adversarial text detection approach is
resilient to white-box attacks. Note that in the above two experiments, the
training accuracy and test accuracy of the BiLSTM are 91.46\%/86.93\% for NA,
80.99\%/78.30\% for RTMR and 87.72\%/87.28\% for IMDB.
% \gl{Note that to obtain sufficient number of benign
% samples, i.e., 1000, we do not limit the length of texts from IMDB in this experiment.} 

\begin{figure}[tp]
  \centering
  \includegraphics[width=0.5\textwidth]{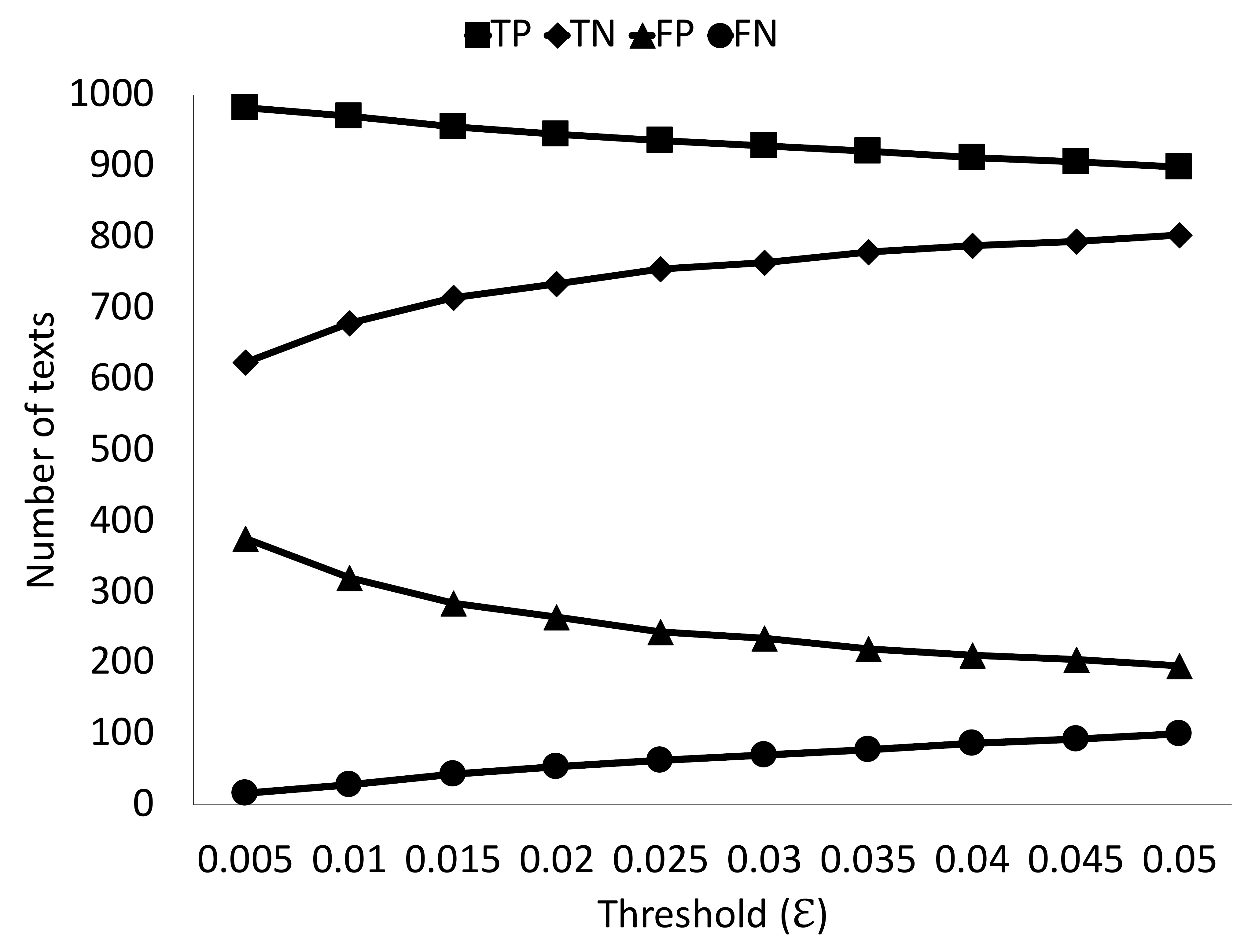}
  \caption{Detection results of our approach with different threshold.The ``TP",``TN'',``FP'' and ``FN'' 
  denote that out of 2000 input texts the number of true positive texts, true negative texts, false positive 
  texts and false negative texts respectively.}
  \label{fig:dtct_eps}
\end{figure}

\textit{Effect of $\epsilon$.} As shown in Algorithm~\ref{alg:df-kl}, the
threshold $\epsilon$ is a key parameter of our detection approach which also has
impact on the follow-up repair. We thus conduct an experiment to exploit the
impact of the threshold on our approach (note that in practice the value of
$\epsilon$ is automatically identified by golden-section search described
before). We adopt TextCNN as the target model and take 1000 adversarial texts
generated by TEXTBUGGER and 1000 normal texts from NA. We vary the threshold
from 0.005 to 0.05 with a step size of 0.005, and record the detection results
with each of the threshold value. The detection results are shown in
Fig~\ref{fig:dtct_eps}. We can observe detection rate gradually decreases (as
the number of true positive decreases) when increasing the threshold and when
the threshold is beyond 0.025, the overall performance of our detection tends to
remain stable. It is as expected that a larger threshold will lead to fewer
false alarms but a higher false negative rate. Our recommendation is to choose a
smaller threshold while keeping the overall performance since the false positive
samples can be mostly mitigated by the following repair which we will show later
in RQ3.

\textit{Effect of Model Architecture.} To exploit the impact of the architecture
of the models used on the proposed algorithm, we introduced another two popular
text classification models which have different architectures: BiLSTM and
FastText~\cite{fasttext}. The FastText is based on a neural network which
incorporates the idea of n-gram for word embedding. To evaluate the performance
of our approach on the two models, for either of them, we first generate 1000
adversarial texts by TEXTBUGGER from NA dataset, and then apply our approach for
detection. The results are shown in the ``Detection results" column of
Table~\ref{tab:ar_im}. We can observe that our approach still can effectively
identify the adversarial samples, i.e. 95\% and 98\% detection rate for BiLSTM and FastText respectively.

% For each of the two models,
% we first generate 1000 adversarial texts by TEXTBUGGER from NA dataset, and then
% apply our detect approach to the adversarial texts. 

\begin{framed}
\emph{Answer to RQ1: Algo.~\ref{alg:df-kl} is effective in detecting adversarial texts with a relatively low false positive rate.} \end{framed}

\noindent \emph{RQ2: Is our hypothesis for voting justified?} To answer this
question, we measure whether the majority of the perturbed texts generated from
an adversarial text have the correct label. We take all the adversarial texts
and apply semantic-preserving perturbations to generate 100 perturbed texts
(using SEAs and TEXTBUGGER) for each of them and measure the percentage of
the perturbed texts that are labeled correctly. The results are in column
``Correctly Labeled'' of Table~\ref{tab:vote}.

\begin{table}[tp]
  \centering
\caption{Results of justifying voting, where ``RP",``SubW'' and ``ParaPer''
    refers to the three semantic-preserving perturbation methods: random
    perturbation, TEXTBUGGER based method and paraphrase based perturbation.}
    
    \small
    % {\small
    % \begin{tabular}{|c|c|c|c|c|c|c|c|c|}
    \begin{tabular}{|c|c|c|c|c|c|}
    \hline
    \multirow{2}{*}{Attack} & \multirow{2}{*}{Dataset} & \multirow{2}{*}{Model} & \multicolumn{3}{c|}{Correctly Labeled (\%)} \\ \cline{4-6}  % & \multicolumn{3}{c|}{Identified as Normal (\%)} \\ \cline{4-9}
    &&  & RP & SubW & ParaPer  \\ \hline % & RP & SubW & ParaPer\\ \hline

    		\multirow{7}{*}{SEAs} & \multirow{2}{*}{NA}   & TextCNN   &68.89	&67.78	&80.16  \\ % &34.35    &68.70  &96.18  \\
           	       							                &     & LSTM      &70.91  &66.96	&79.86  \\ % &37.67    &76.71  &91.82   \\
                     		     & \multirow{2}{*}{RTMR}  & TextCNN   &80.21 	&78.45	&89.80  \\ % &64.00    &77.33  &98.00  \\
                					       		            &     & LSTM      &65.66	&65.09	&84.85  \\ % &74.44    &79.70  &99.25  \\
        					           & \multirow{2}{*}{IMDB}  & TextCNN   &87.64	&79.07	&86.51  \\ % &69.53    &67.19  &98.44   \\
                  						                  &     & LSTM      &52.63 	&55.00	&80.99  \\ \cline{2-6} % &62.30     &65.57  &99.18 \\ \cline{2-9}
                   &\multicolumn{2}{c|}{Avg}	&\textbf{70.99} &\textbf{68.73}  &\textbf{83.70} \\ \hline % &\textbf{57.05}  &\textbf{72.53}   &\textbf{97.14}  \\ \hline

\multirow{7}{*}{TextB} & \multirow{2}{*}{NA}    & TextCNN        &91.82 &81.10	&83.66 \\ %&71.43  &82.47  &100.00 \\
                 						 		            &        & LSTM      &88.24	&79.85	&89.29 \\ %&60.00  &78.82  &99.41\\
          					         & \multirow{2}{*}{RTMR} & TextCNN   &86.36	&80.41  &82.56 \\ %&88.00  &84.57  &98.29\\
                          					         &       & LSTM      &79.64	&69.88 	&80.22 \\ %&91.26  &90.71  &99.45  \\
                    			   & \multirow{2}{*}{IMDB} & TextCNN   &97.84	&90.80	&91.09 \\ %&91.58  &80.69  &100.00  \\
                         					           &       & LSTM      &84.53	&80.57	&91.63 \\ \cline{2-6} %&72.11  &69.72  &100.00 \\  \cline{2-9}
&\multicolumn{2}{c|}{Avg}	&\textbf{88.07}	&\textbf{80.44}		&\textbf{86.41}  \\  \hline %&\textbf{79.06} &\textbf{81.16} &\textbf{99.53} \\ \hline	

\multirow{7}{*}{TextF} & \multirow{2}{*}{NA}         & TextCNN   &38.00  &52.95	 &78.26       \\ % &54.29  &62.34  &99.45         \\ 
                 						 		            &        & LSTM      &40.21	 &65.25	 &84.47       \\ % &28.09  &68.58  &98.79          \\
          					         & \multirow{2}{*}{RTMR} & TextCNN   &59.90	 &57.96  &84.10       \\ % &51.42  &69.94  &91.81         \\
                          					         &       & LSTM      &62.45	 &55.47  &76.49       \\ % &53.48  &67.30  &92.69       \\ 
                    			   & \multirow{2}{*}{IMDB} & TextCNN   &82.50	 &78.38	 &95.10       \\ % &68.12  &71.86  &94.55        \\   
                         					           &       & LSTM      &59.20	 &66.84	 &94.44       \\  \cline{2-6} % &62.41  &67.74  &91.6       \\ \cline{2-9}
                             &\multicolumn{2}{c|}{Avg}	&\textbf{57.04}	&\textbf{62.80}	&\textbf{85.48}    \\  \hline % &\textbf{52.97}	&\textbf{67.96}		&\textbf{94.82}  \\ \hline	

                             \multicolumn{3}{|c|}{Avg} &\textbf{72.03}	&\textbf{70.66}		&\textbf{85.20}  \\ \hline %  &\textbf{63.02}  &\textbf{73.88} &\textbf{97.16} \\ \hline
					
\end{tabular}
\label{tab:vote}
\end{table}

We can observe that our hypothesis holds across all models, methods used to
generate adversarial texts and perturbation methods, i.e., the percentage of
perturbed texts with correct labels is more than 50\% in all cases. Comparing
the results on different perturbation methods, perturbation using ParaPer
restores the correct label significantly more often than the other two. This is
expected since the ParaPer is paraphrase-based, which preserves the most
semantics when generating adversarial texts among the three methods. Comparing
different adversarial texts, adversarial texts generated by TEXTBUGGER, once
perturbed, are more likely to have the correct label than those generated by
SEAs and TEXTFOOLER. This is reasonable as the adversarial texts generated by
SEAs and TEXTFOOLER are more semantically similar to the original texts compared
with these texts generated by TEXTBUGGER.

% Note that in Algorithm~\ref{alg:repair}, only perturbed texts which are
% identified as `normal' by Algorithm~\ref{alg:df-kl} are allowed to vote. We thus
% further evaluate how likely a perturbed text is identified as `normal'. This is
% relevant as otherwise it may take a longer time to collect enough votes. The
% results are shown in column `Identified as Normal' of Table~\ref{tab:vote}. It
% can be observed that most of the perturbed texts are identified as normal and
% thus allowed to vote. Comparing different perturbations, almost all texts
% generated using ParaPer are identified as normal.

\begin{framed}
\emph{Answer to  RQ2: Our hypothesis for voting is justified.}
\end{framed}

\begin{table*}[t]
    \centering
    \caption{Overall repair accuracy (\%) comparison between our approach and two baselines.}
    % \small
    \label{tab:overallR}
      \begin{tabular}{|c|c|c|c|c|c|c|c|}
      \hline
      \multirow{2}{*}{Attack} & \multirow{2}{*}{Dataset} & \multirow{2}{*}{Model} & \multicolumn{3}{c|}{Our Approach} & \multicolumn{2}{c|}{Baselines} \\
  \cline{4-8}          &       &       & RP    & SubW  & ParaPer & Autocorrect & scRNN \\
      \hline
      \multirow{6}{*}{SEAs} & \multirow{2}{*}{NA} & TextCNN   & 23.66  & 42.75  & 70.23  & 4.70  & 27.52  \\ \cline{3-8}
                       &       & LSTM  & 26.71  & 49.32  & 67.81  & 5.66  & 28.93  \\ \cline{2-8} 
                       & \multirow{2}{*}{MR} & TextCNN   & 50.00  & 60.00  & 76.00  & 6.76  & 30.18  \\ \cline{3-8}
                       &       & LSTM  & 50.38  & 51.13  & 78.20  & 5.71  & 19.05  \\ \cline{2-8}
                       & \multirow{2}{*}{IMDB} & TextCNN   & 60.94  & 53.13  & 79.69  & 4.82  & 38.69  \\  \cline{3-8}
                       &       & LSTM  & 35.25  & 34.42  & 75.41  & 10.12  & 34.34  \\ \cline{2-8}
                       &\multicolumn{2}{c|}{Avg} & \textbf{41.16} & 	\textbf{48.46} &	\textbf{74.56} &	\textbf{6.30} &	\textbf{29.79}   \\                       
      \hline
  \multirow{6}{*}{TEXTBUGGER} &\multirow{2}{*}{NA} & TextCNN   & 64.94  & 67.53  & 79.74  & 18.79  & 29.09  \\
  \cline{3-8}          &       & LSTM  & 52.35  & 58.82  & 82.84  & 18.79  & 36.81  \\
  \cline{2-8}          &\multirow{2}{*}{MR} & TextCNN   & 72.57  & 67.43  & 80.01  & 19.21  & 29.26  \\
  \cline{3-8}          &       & LSTM  & 69.40  & 59.02  & 79.23  & 23.81  & 25.97  \\
  \cline{2-8}          &\multirow{2}{*}{IMDB} & TextCNN   & 87.62  & 73.27  & 93.56  & 27.19  & 68.86  \\
  \cline{3-8}          &       & LSTM  & 62.15  & 56.57  & 90.44  & 29.63  & 64.81  \\
  \cline{2-8}          &\multicolumn{2}{c|}{Avg} & \textbf{68.17} &	\textbf{63.77} &	\textbf{84.30} &	\textbf{22.90} &\textbf{42.47 } \\  
  \hline
  
  \multirow{6}{*}{TEXTFOOLER} &\multirow{2}{*}{NA}  & TextCNN   &37.37   &52.00   &70.71    &12.00   &22.60   \\
  \cline{3-8}                               &       & LSTM      &38.21   &46.21   &75.51    &12.90   &21.60   \\
  \cline{2-8}                 &\multirow{2}{*}{MR} & TextCNN   &40.39   &48.21   &54.95    &20.50   &23.30   \\
  \cline{3-8}                               &       & LSTM      &39.35   &41.49   &56.99    &22.10   &23.50   \\
  \cline{2-8}                 &\multirow{2}{*}{IMDB} & TextCNN   &82.35   &71.57   &90.69    &39.30   &57.21   \\
  \cline{3-8}                               &       & LSTM      &53.97   &45.63   &87.30    &36.07   &57.14   \\
  \cline{2-8}                 &\multicolumn{2}{c|}{Avg} &\textbf{48.61}   &\textbf{50.85}   &\textbf{72.69}   &\textbf{23.81}   &\textbf{34.23}  \\  
  \hline

  \multicolumn{3}{|c|}{Avg} &\textbf{52.65}   &\textbf{54.36}   &\textbf{77.18}   &\textbf{17.67}   &\textbf{35.49}    \\
  \hline
  \end{tabular}%
\end{table*}%

\begin{table}[tp]
  \centering
  \caption{Example of adversarial text repaired with our approaches.}
  \begin{tabular}{|l|l|}
  \hline
  \multirow{2}[1]{*}{Ori} &  \textit{a silly, self-indulgent film about a silly,}\\
    & \textit{self-indulgent filmmaker}\\ \hline
  Adv&  \textit{a silly, indulgent film about a silly, indulgent director}  \\ \hline
  RP & \textit{a silly, decadent films about a silly, indulgent director}\\  \hline
  SubW & \textit{a silly, sumptuous movies about a silly, indulgent director}\\  \hline
  ParaPer & \textit{a silly, sumptuous movie about a silly, indulgent director}\\ \hline
  \end{tabular}
  \label{tab:exmpf}
\end{table}

\noindent \emph{RQ3: Is our approach effective in repairing adversarial texts?}
To answer this question, we systematically apply Algorithm~\ref{alg:repair} to
all the adversarial texts, and measure its overall repair accuracy. Formally,
the overall repair accuracy is defined as follows:
\begin{align}
  \textit{overall repair accuracy} = \frac{\textit{\#correctly repaired texts}}{\textit{\#adversarial texts}}
\end{align}
where \textit{\#adversarial texts} denotes the total number of adversarial texts
to repair, and $\textit{\#correctly repaired texts}$ denotes the number of texts
which can be correctly predicted after repair. We set the parameters for SPRT
as follows: the error bounds $\alpha$ and $\beta$ are both set as $0.1$, the
confidence threshold $\rho$ is $0.8$, and the indifference region $\sigma$ is to
be $0.2\times\rho$. We remark that a higher confidence can be achieved by
setting a larger threshold and smaller error bounds. The price to pay is that it
would typically require more perturbed texts (and thus time overhead). Note that
when we apply ParaPer to generate perturbed texts, we use 25 different target
languages for generating 25 semantic-preserving perturbations through two
translations. If more is required, we use two target languages each time (and
three translations), which provides us additionally $25\times 25$ perturbed
texts. To be consistent with ParaPer, we set the perturbation budget (maximum
number of perturbations) for RP and SubW as 650 as well. 

We compare our approach with two baselines~\cite{gao2018black,pruthi2019combating}. Both baselines can
automatically detect and correct adversarial examples with misspellings. The
first baseline~\cite{gao2018black} used the Python autocorrect
package~\footnote{https://pypi.org/project/autocorrect/} to detect and
automatically correct the adversarial texts with misspellings. In the following,
we refer this baseline to Autocorrect. The second
baseline~\cite{pruthi2019combating} proposed a word recognition model scRNN for
the same task. We first attempt to repair adversarial texts using each
baseline and then test the accuracy of the target model on the repaired texts.

We summarize the results of different models and datasets in
Table~\ref{tab:overallR}. On average, we are able to correctly repair 54.66\%,
56.12\% and 79.43\% of the adversarial texts using RP, SubW and ParaPer
respectively, while the two baselines achieve 14.60\% and 36.91\%. That is, all
the three sort of methods in our approach outperform the two baselines and
ParaPer achieves the best overall performance among the three. Comparing
adversarial texts generated using different methods, we observe that adversarial
texts generated by SEAs are harder to repair than those generated by TEXTBUGGER.
This is expected as adversarial texts generated by SEAs (with two translation)
are often structurally different from the original normal texts, whereas
adversarial texts generated by TEXTBUGGER are very similar to the original
normal texts. Comparing different repairing methods on different adversarial
texts, we see that ParaPer performs significantly better than the other methods.
This is expected to be due to the same reason above, i.e., other methods are
ineffective in repairing adversarial texts which are structurally different from
the original normal texts. The performance of the two baselines are
significantly worse than our approaches, i.e., at least 17.75\% gap (between the
best performance of baselines and the worst performance of our approach), which
is as expected since the two baselines are to detect the misspellings and thus
are not able to handle semantic-preserved adversarial texts. Surprisingly, the
performance of RP is close to that of SubW on adversarial texts generated
TEXTBUGGER. The possible explanation is that these adversarial texts are near to
the classification boundary and thus a random perturbation is sufficient for the
repair. Table~\ref{tab:exmpf} shows a concrete example of repaired text using
different perturbation methods. The first row shows an original text from RTMR,
the label of which is negative. The adversarial text, at the second row, is
generated by TEXTBUGGER with the LSTM model. The subsequent rows then show
successful repairs which are generated using different methods. Note that by
suggesting a simple edit (of one word), the text is no long considered
adversarial and would be labeled correctly using the trained model.

We also compare our approach with the adversarial training method. We retrained
the target model by adding 10\% of adversarial texts (half of them are generated
by TEXTBUGGER and half by TEXTFOOLER) into the training set. The retraining
procedure is stopped once its accuracy on test set reaches the original level
and  at least 90\% of adversarial texts in the training set can be correctly
predicted. We compare the performance of the two methods from the following two
aspects. Firstly, we compare the robustness of the models obtained through the
two approaches. The results are shown in Table 13. We can observe that,
respectively, 78.5\% and 51.92\% of adversarial texts can be predicted correctly
by our approach and models from adversarial training. Secondly, we conducted
experiments to evaluate if the model obtained through adversarial training is
robust against different attacks. The results are shown in Table 14. We can
observe that the success rate of attacking indeed decreases, but not
significantly, i.e. a 3.3\% drop on average. This is consistent with the
well-known result that adversarial training easily overfits and has limited
effectiveness in defending against unknown attacks~\cite{zhang2019limitations},
which is also evidenced in \cite{textfooler} where adversarial training only
decreases the attack success rate by 7.2\% on MR dataset. On the other hand, our
approach is resilient under different kinds of attacks with a totally different
defense paradigm, i.e. decreasing the attack success rate by 59.4\% on average.

\begin{table}[tp]
  \centering
  \caption{The accuracy of adversarial-training models against adversarial texts. ``Acc1" denotes the accuracy of the adversarially retrained models, ``Acc2" denotes the accuracy of our approach (using ParaPer).}
    \begin{tabular}{|c|c|c|c|c|}
    \hline
    {Attack} & {Dataset} & {Model} & {Acc1(\%)} & { Acc2(\%)}  \\
    \hline
    \multirow{7}{*}{TextB} & \multirow{2}{*}{NA}   & {CNN}  & 39.8  &79.74      \\ \cline{3-5}
                                        &          & {LSTM} & 61.3  &82.84      \\ \cline{2-5}          
                          & \multirow{2}{*}{RTMR}  & {CNN}  & 30.3  &80.01      \\ \cline{3-5}          
                                          &        & {LSTM} & 47.1  &79.23      \\ \cline{2-5}
                          & \multirow{2}{*}{IMDB}  & {CNN}  & 44.63 &93.56      \\ \cline{3-5}
                                           &       & {LSTM} & 70.04 &90.44      \\ \cline{2-5}
                        & \multicolumn{2}{c|}{Avg} & \textbf{48.86} & \textbf{84.30 } \\ \hline
    \multirow{7}{*}{TextF}  & \multirow{2}{*}{NA}   & {CNN}  & 60.6  &70.71     \\ \cline{3-5}
                                         &          & {LSTM} & 68.1  &75.51     \\ \cline{2-5}
                            & \multirow{2}{*}{RTMR} & {CNN}  & 31.9  &54.95     \\ \cline{3-5}
                                            &       & {LSTM} & 46.9  &56.99     \\ \cline{2-5}
                            & \multirow{2}{*}{IMDB} & {CNN}  & 56.33 &90.69     \\ \cline{3-5}
                                            &       & {LSTM} & 66.07 &87.3     \\ \cline{2-5}
                            & \multicolumn{2}{c|}{Avg} & \textbf{54.98}   &\textbf{72.69}\\   \hline
    \multicolumn{3}{|c|}{Avg} &\textbf{51.92}    & \textbf{78.5} \\   \hline          
    \end{tabular}%
  \label{tab:adv-train}%
\end{table}%

% \usepackage{multirow}
\begin{table}[tp]
  \centering
  \caption{Success rate(\%) of attacking different models (with TEXTBUGGER). The column ``advR"
  refers the attack success rate on the adversarially retrained model, the
  column ``Ori" refers the attack success rate on original model, and the last column ``Ours'' is the
  attack success rate of attacking our approach.}
  \begin{tabular}{|c|c|c|c|c|c|c|c|}
  \hline
  \multirow{2}{*}{Dataset} & \multicolumn{3}{c|}{advR} & \multicolumn{3}{c|}{Ori} & \multirow{2}{*}{Ours} \\ \cline{2-7}
         & CNN  & LSTM & Avg   & CNN  & LSTM & Avg     &  \\ \hline
  NA     & 49.4 & 52.8 & \textbf{51.1}  &53.1  &63.3  &\textbf{58.2}    & \textbf{1.6 }          \\ \hline
  RTMR   & 71.7 & 77.8 & \textbf{74.8}  &74.1  &78.4  &\textbf{76.3}    & \textbf{28.4} \\ \hline
  IMDB   & 62.5 & 95   & \textbf{78.8}  &63.9  &96.3  &\textbf{80.1}    & \textbf{6.5 }\\ \hline
  \end{tabular}
\end{table}

% \begin{table}[t]
%   \centering
%   \caption{Success rate(\%) of attacking different models. The column
%   ``Retrained model" refers the success rate of attacking the adversarially
%   retrained model and attacking the model retrained without adversarial texts.}
%     \begin{tabular}{|c|c|c|c|}
%     \hline
%     {Dataset} & {Model} & {Retrained model} & {Our approach}  \\
%     \hline
%                        \multirow{2}{*}{NA} & CNN   & 49.4 & 53.1   \\ \cline{2-4}
%                                            & LSTM  & 52.8 & 63.3    \\ \cline{2-4}
%                                            & Avg   & \textbf{51.1}   &\textbf{58.2}  \\

%     \hline
%     \multirow{2}{*}{RTMR}                 & CNN   & 71.7          & 74.1    \\ \cline{2-4}
%                                           & LSTM  & 77.8          & 78.4    \\ \cline{2-4}
%                                           & Avg   &\textbf{74.75}	&\textbf{76.25} \\
%     \hline
%     \multirow{2}{*}{IMDB}                 &CNN    &66.5           & 63.9    \\ \cline{2-4}
%                                           &LSTM   &95             & 96.3    \\ \cline{2-4}
%                                           &Avg    &\textbf{80.75}	& \textbf{80.1} \\
%     \hline
%     \end{tabular}
%   \label{tab:retrain-attack}
% \end{table}

% We add
% adversarial texts (10\% of the training set) into the training set and retrain
% the target model until at leat 90\% of the adversarial texts in the training set can be correctly predicted.
% randomly select 10\% of the
% training set and 

\textit{Effectiveness on False Positive Samples}. Considering that our detection
approach may report false positive samples, one question is whether our repair
is effective on these samples. To address this concern, we conduct a simple
experiment on NA dataset with TextCNN and LSTM. Concretely, we apply our
approach to repair randomly selected 1000 samples which are wrongly detected as
adversarial. The results show that 81.4\% (for TextCNN) and 85.2\% (for LSTM) of
samples can be correctly classified after repair. This suggests that our
approach can correctly handle most of the false positive samples. It is
reasonable that our approach can repair the false positive samples effectively
as these false positive samples are mostly wrongly identified because of the
large KL divergence (though their final predicted labels could be the same). As a
result, we only need to generate repaired candidates with a KL divergence
smaller than the threshold, which could be achieved effectively with Algo.\ref{alg:tbp}.

% we first take one of the two
% models, e.g., TextCNN, as the target model, and then from the test set, we
% select all the samples which can be correctly classified by the target model.
% After that, we use our detection method, i.e., Algorithm~\ref{alg:df-kl}, to
% randomly select 1000 false positive samples and use Algorithm~\ref{alg:repair}
% (we employ SubW to perform the semantic preserving perturbation) to repair them.

\textit{Effectiveness on a Third Model.} We also exploit the effectiveness of
our approach in the case where the adversarial texts from a model which is
different from the two models used in detection. The results are shown in the
column ``Repair accuracy" of Table~\ref{tab:taa}. We can observe that our
approach achieves 55.83\% repair accuracy on average, which suggests that our
approach is still effective in handling this sort of adversarial text. In
general, adversarial samples from the third models can be categorized into two
groups: 1) adversarial samples which are invalid for both of our models (used
for adversarial detection), and 2) adversarial samples which can still fool at
least one of our models. In the first case, even if it is wrongly identified as
adversarial (because of the large KL divergence), our approach can still produce
a right prediction with high probability (see "Effectiveness on False Positive
Samples" above). In the second case, our approach is able to repair the
adversarial sample just in the same way with adversarial samples from our own
models.

\begin{figure}[t]
  \centering
  \includegraphics[width=0.5\textwidth]{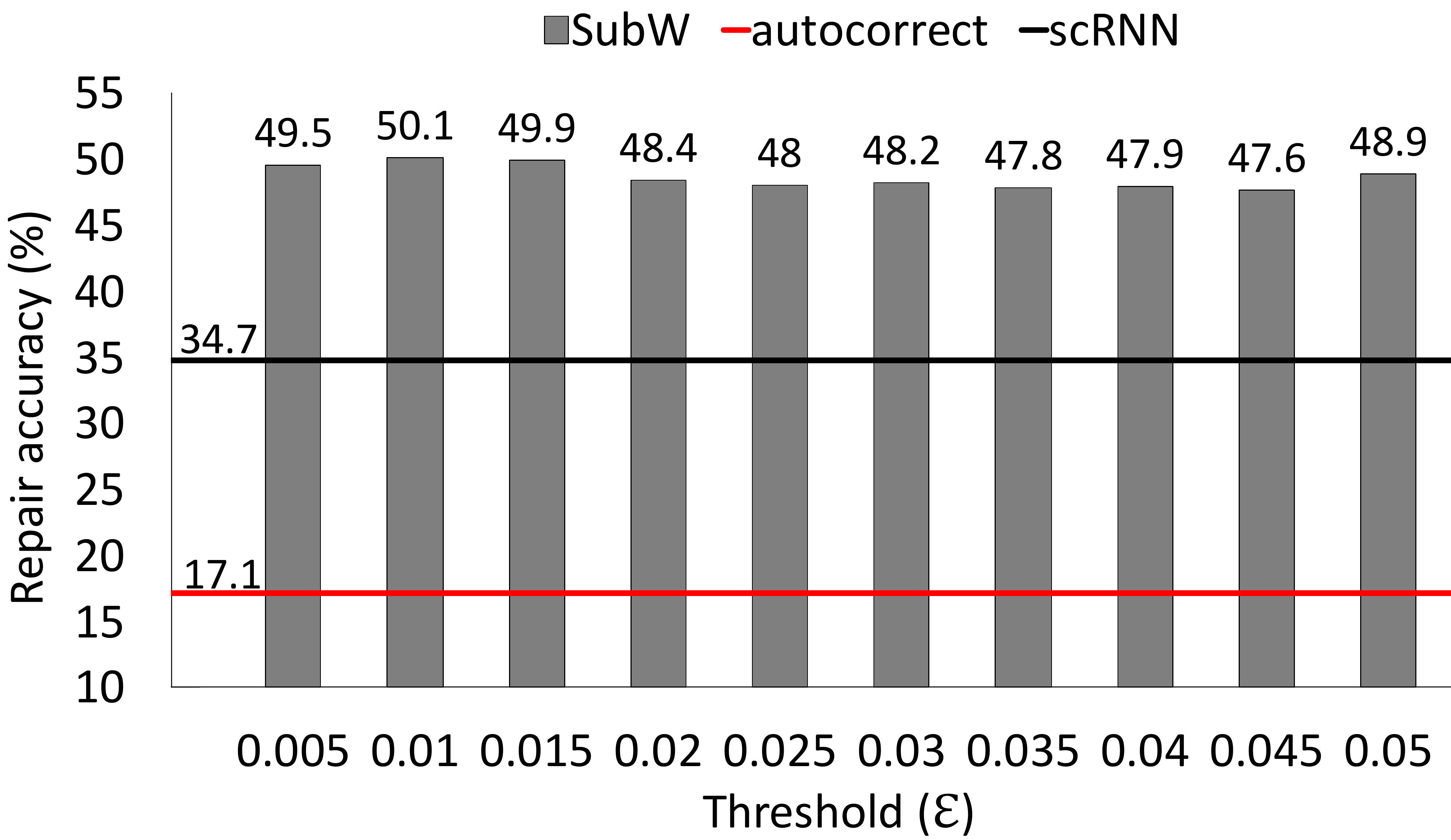}
  \caption{Repair accuracy of our approach with different threshold. The two solid lines denotes the 
  repair accuracy of two baselines, i.e., autocorrect (red line) and scRNN (black line)}
  \label{fig:repair_eps}
\end{figure}

\textit{Effect of $\epsilon$.} As mentioned in RQ1, the threshold $\epsilon$
also has impact on the repair approach. Thus, following the setting of
Fig~\ref{fig:dtct_eps}, we adopt the SubW perturbation method and show the
impact of threshold on the repair part of our approach. Note that the repair is
only conducted on the 1000 adversarial texts. The results are shown in
Fig~\ref{fig:repair_eps}. We can observe that the repair accuracy gradually
drops in general, but still remains stable and significantly more effective
compared with the two baselines. The possible reason is that: essentially
our approach is effective mainly depending on the voting mechanism which blurs
the effect of $\epsilon$, while the two baselines repair the adversarial texts
only by correcting misspellings (or replacing an unknown word to a possible
one), as explained before, which has limited capability to deal with the
semantic-preserved adversarial texts.

% Meanwhile, $\epsilon$ is mainly used to
% improve the accuracy in the detection phase of our approach, while in the phase
% of repairing, it is mainly used to select generated texts which are likely to be
% ``normal'' sample. The selection do noAccording to Table~\ref{tab:vote},
% The relatively stable results can be explained as that
% $\epsilon$ is mainly used to improve the accuracy in the detection part of our
% approach, while when repairing, our approach mainly depends on voting and the
% $\epsilon$ is mainly used to select the generated repairs. According to Table~\ref{tab:vote}, without the selection, the majority voting 

\textit{Effect of Model Architecture.} We also show the impact of architecture
of models on our repair approach. We take the two models again, i.e., BiLSTM and
FastText, and apply our repair approach to 1000 adversarial texts for each
model. The repair results are shown in the column ``Repair accuracy'' of
Table~\ref{tab:ar_im}. We can observe our approach achieves 57.8\% and 74\%
overall repair accuracy for BiLSTM and FastText respectively, which, again,
shows the robustness of our approach’s performance on different architectures.
We notice that the average repair accuracy is even better compared with TextCNN
and LSTM. This is reasonable as an adversarial text is repaired by our approach
using adversarial perturbation methods. As a result, a more easily attacked
model is likely to be more easily repaired. In our experiments, the attack
success rate of TextCNN on NA is 55\% while that of FastText is 78.34\%, and
correspondingly the repair accuracy of TextCNN is lower than that of FastText,
i.e., 67.53\% and 74\% respectively. We also notice that the repair accuracy of
BiLSTM and LSTM are close (i.e., 57.8\% and 58.82\% respectively) as the attack
success rate of the two models are comparable (i.e., 62.9\% and 60.67\%
respectively).

\begin{framed}
\emph{Answer to RQ3: Our approach can repair about 80\% of the adversarial texts. ParaPer performs the best.}
\end{framed}

\begin{table*}[t]
  \centering
  \caption{Time overhead }
  {
  \small 
  \begin{tabular}{|c|c|c|c|c|c|c|}
    \hline
    \multirow{2}{*}{Attack} & \multirow{2}{*}{Dataset} & \multirow{2}{*}{Model} & \multirow{2}{*}{\tabincell{c}{ Detect (ms)} }  & \multicolumn{3}{c|}{Repair (s)} \\ \cline{5-7}
    &&  &  & RP & SubW & ParaPer\\ \hline
    \multirow{7}{*}{SEAs} & \multirow{2}{*}{NA} & TextCNN   &7.6 	&55.1	&1.2	&181.6  \\
                  				            &         & LSTM      &8.2 	&48.8	&1.0	&223.2   \\
                       & \multirow{2}{*}{RTMR} & TextCNN    &3.2	&46.3	&1.4	&171.6  \\
                				            	&        & LSTM       &3.3 	&36.5	&1.1	&144.0  \\
        			        & \multirow{2}{*}{IMDB}  & TextCNN    &13.4 &61.5	&1.0	&134.8  \\
                  			             	&       & LSTM        &13.3 &92.7 &1.1	&167.3  \\ \cline{2-7}
                                  &\multicolumn{2}{c|}{Avg}&\textbf{8.2}&\textbf{56.8}&\textbf{1.1}&\textbf{170.5} \\ \hline

  \multirow{7}{*}{TEXTBUGGER} & \multirow{2}{*}{NA} & TextCNN   &6.6  	&35.0	  &0.7 	&157.8   \\
                  				                      &   & LSTM      &6.4 	  &31.2 	&0.8	&171.5   \\
          					      & \multirow{2}{*}{RTMR}   & TextCNN   &3.6 	  &39.6	  &1.4	&81.1 \\
                                            &       & LSTM      &3.4    &40.0	  &1.2	&67.7\\
                    			& \multirow{2}{*}{IMDB}   & TextCNN   &15.2   &51.7   &0.8  &47  \\
                          	&                       & LSTM      &17.3   &101.0	&1.0	&76.3   \\
     \cline{2-7}
     &\multicolumn{2}{c|}{Avg}	     & \textbf{8.8}&\textbf{49.7}&\textbf{1.0}&\textbf{102.8} \\ \hline

  \multirow{7}{*}{TEXTFOOLER} & \multirow{2}{*}{NA} & TextCNN   &6.7 	 &47.4	 &0.7  & 109.0  \\
                                                &   & LSTM      &8.6	 &54.5	 &0.7  & 98.0   \\
                            & \multirow{2}{*}{RTMR} & TextCNN   &4.9   &34.2   &0.8	 & 85.7   \\
                                            &       & LSTM      &4.8   &46.9   &1.1	 & 90.3  \\
                            & \multirow{2}{*}{IMDB} & TextCNN   &28.9  &141.4  &1.5   & 79.4  \\
                                            &       & LSTM      &32.8  &102.6	 &1.1	 & 99.6 \\ \cline{2-7}

&\multicolumn{2}{c|}{Avg}	  & \textbf{14.5}&\textbf{71.2}&\textbf{1.0}&\textbf{93.7} \\ \hline

     \multicolumn{3}{|c|}{Avg}   &\textbf{10.5}&\textbf{59.2}&\textbf{1.0}&\textbf{122.4} \\ \hline \end{tabular}
}
\label{tab:avg-costs}
\end{table*}

\noindent \emph{RQ5: What is the time overhead of our approach?} The time
overhead of our approach mainly consists of two parts: detection and repairing.
For detection, measuring the time spent is straightforward. For repairing,
precisely measuring the time is a bit complicated. For RP and SubW, the time
taken to obtain the synonymy might be different depending on the configuration
of \textit{gensim}. For ParaPer, our implementation uses an online NMT service
which often suffers from network delay and as a result, the time measure is
inaccurate. To discount the effect of the network delay, we thus count the
average number of perturbed texts required for voting, which is then multiplied
with the average time needed to obtain a perturbed text using the respective
methods. According to our empirical study on 1000 trials, the average
time taken for generating one perturbed text is  0.55 seconds for RP, 0.09
seconds using SubW, and 1.44 seconds for ParaPer.

The results are summarized in Table~\ref{tab:avg-costs} where column `Detect' is
the average detection time and column `Repair' is the average repair time. The
results show that detection is very efficient, i.e., the maximum time used is
17.3 ms and the average time across all datasets are 8.2 ms, 8.8 ms and 14.5
ms for SEAs, TEXTBUGGER and TEXTFOOLER generated adversarial texts respectively. This is
expected as Algorithm~\ref{alg:df-kl} only requires to obtain the probability
vectors of two neural network models and compare their difference to a
threshold. Note that the more complex a model is, the more time is required. For
example, detecting adversarial texts from IMDB requires more time than those
from RTMR as the IMDB models are more complex.

For repairing, RP needs 59.2 seconds on average (maximum 141 seconds); SubW
needs 1 seconds on average (maximum 1.5 seconds); ParaPer needs 122.4 seconds on
average (maximum 223.2 seconds. Repairing using SubW takes much less time as
SubW is designed to generate perturbed texts under the guidance of $D_{KL}$ and
the resulting texts thus have a much higher probability to be detected as
normal. Besides, we observe that repairing adversarial texts generated by SEAs
and TEXTFOOLER are more difficult (consistent with the above). On average, the
time needed for repairing adversarial texts generated by the three methods are
76.13 seconds, 51.16 seconds and 55.3 seconds respectively. The results show that
adversarial texts generated by TEXTFOOLER are relatively time-consuming to be
repaired compared with that of TEXTBUGGER. This is reasonable since the adversarial texts generated by TEXTFOOLER is more nature
compared with that of TEXTBUGGER. If our approach is to be used in an online
setting, we thus would recommend repairing with SubW which repairs 77\% of the
adversarial texts with a total time overhead of 1.1 seconds. We remark that we
can easily parallelize the generation of perturbed texts to reduce the time
overhead for all three methods.

\begin{framed}
\emph{Answer to  RQ4: Our approach has the potential to detect and repair adversarial texts at runtime. }
\end{framed}

\subsection{Threats to Validity}

\noindent \textit{Quality of NMTs} Our SEAs perturbation method requires the availability of multiple NMTs. In this work, we utilize the online industrial NMTs. The quality of NMTs will influence the performance of our repair algorithm, i.e., we might need more perturbations for a successful repair with worse NMTs.

\vspace{1mm}
\noindent \textit{Word substitution} Both random perturbation and Algorithm~\ref{alg:tbp} work by replacing selected words with their synonyms. Currently, we look for synonyms by searching the neighborhood of a given text in the embedding space. However, this may not always find the ideal synonyms, i.e., words which cause syntactical or grammar errors may be returned.
Besides, finding better synonyms usually takes more time, which can be time-consuming.

\vspace{1mm}
\noindent \textit{Limited datasets and adversarial texts} Our experiments results are subject to the selected datasets and generated adversarial texts, which have a limited number of labels. In general, it is difficult to vote for the correct label if there are many candidate labels, i.e., more perturbations are needed. Besides, we evaluate our approach on two existing attacks, it is not clear if our algorithm repairs adversarial texts from future attacks.

\section{Related works}
\label{sec:re}
This work is related to work on adversarial attacks in the text domain, which can be roughly divided into the following categories. One category is adversarial misspelling, which tries to evade the classifier by some ``human-imperceptible'' misspelling on certain selected characters~\cite{hotflip,ebrahimi2018adversarial,textbugger}. The core idea is to design a strategy to identify the important positions and afterwards some standard character-level operations like insertion, deletion, substitution and swap can be applied.
Another category is adversarial paraphrasing. Compared to misspelling, paraphrasing aims to generate semantics-preserving adversarial samples either by replacing certain words with their synonyms~\cite{textbugger} or paraphrasing the whole sentence~\cite{sears,scpns}. For instance, the work in~\cite{sears} uses NMTs to paraphrase the input; in work~\cite{scpns}, the authors proposed Syntactically Controlled Paraphrase (SCPNs) to generate adversarial texts with the desired syntax. Our work uses paraphrasing as a way of generating repairs instead.

This work is related to detect adversarial perturbation. Existing detection methods for adversarial perturbation mainly focuses on the image domain~\cite{zheng2018robust,liu2019detection,xu2019adversarial,wang2019adversarial}. Recently, Rosenberg \emph{et al.} devised a method to detect adversarial texts for Recurrent Neural Networks~\cite{rosenberg2019defense}. The idea is to compare the confidence scores of the original input and its squeezed variant. An input is regarded as adversarial if the two confidence scores are significantly different. 

This work is related to work on defending adversarial perturbation, which mainly focus on the image domain, e.g., adversarial training~\cite{goodfellow2014explaining,tramer2017ensemble} and robust optimization\cite{madry2017towards}.
Rosenberg \emph{et al.}~\cite{rosenberg2019defense} presented several defense methods for adversarial texts, like adversarial training in the text domain or training ensemble models. Pruthi \emph{et al.}~\cite{pruthi2019combating} proposed to place an auxiliary model before the classifier. The auxiliary model is separately trained to recognize and correct the adversarial spelling mistakes. In~\cite{wang2019natural}, Wang \emph{et al.} proposed \textit{Synonyms Encoding Method} to defend adversarial texts in the word level, which maps all the semantically similar words into a single word randomly selected from the synonyms. The approach is shown to be effective to resist attacks generated by word-substitution.

To the best of our knowledge, our approach is the first to repair the adversarial texts without modifying/retraining the model and thus is complementary to existing approaches.

\section{conclusion}
\label{sec:con}
In this work, we propose an approach to automatically detect and repair
adversarial texts for neural network models. Given an input text to a pair of
neural network models, we first identify whether the input is adversarial or
normal. Afterwards, we automatically repair the adversarial inputs by generating
semantic-preserving perturbations which collectively vote for the correct label
until a consensus is reached (with certain error bounds). Our experiments on
multiple real-world datasets show the effectiveness of our approach.
 
%  \section*{Acknowledgements}
% This research is supported by the National Key R\&D Program of China under Grant No. 2019YFB1600700  and Project of Science and Technology Research and Development Program of China RailwayCorporation (P2018X002) and NSFC Program (Grant No. 61972339). This research is also supported by the National Research Foundation, Singapore under its AI Singapore Programme (AISG Award No: AISG-RP-2019-012). Any opinions, findings and conclusions or recommendations expressed in this material are those of the author(s) and do not reflect the views of National Research Foundation, Singapore. This research has also been supported by the Key-Area Research and Development Program of Guangdong Province (Grant no. 2018B010107004), the Fundamental Research Funds for the Zhejiang University NGICS Platform, and the National Research Foundation, Prime Minister’s Office, Singapore under its Corporate Laboratory@University Scheme, National University of Singapore, and Singapore Telecommunications Ltd.
% 
\bibliographystyle{IEEEtran}
\bibliography{ref}
\end{document}